\documentclass[11pt]{article}
\usepackage[preprint]{acl}
\usepackage{times}
\usepackage{latexsym}
\usepackage[T1]{fontenc}
\usepackage[utf8]{inputenc}
\usepackage{microtype}
\usepackage{inconsolata}
\usepackage{graphicx}

\usepackage{tikz}
\usetikzlibrary{plotmarks}
\usepackage{pgfplots}
\usepackage{pgfplotstable}
\usepgfplotslibrary{groupplots}
\usepgfplotslibrary{colormaps}
\pgfplotsset{compat=1.18}

\usepackage[x11names, table]{xcolor}
\usepackage{booktabs}
\usepackage{multicol}
\usepackage{multirow}
\usepackage{makecell}
\usepackage{enumitem}
\usepackage{amsmath}
\usepackage{amssymb}
\usepackage{cleveref}
\usepackage{twemojis}
\usepackage{xurl}

\usepackage[htt]{hyphenat}
\usepackage{silence}
\usepackage{gradient-text}
\newcommand{\rainbow}[1]{\gradientRGB{#1}{255,0,0}{150,0,250}}

\newcommand\footnotenonumber[1]{%
  \begingroup
  \renewcommand\thefootnote{}\footnote{#1}%
  \addtocounter{footnote}{-1}%
  \endgroup
}

\definecolor{oiOrange}{HTML}{E69F00}
\definecolor{oiSkyBlue}{HTML}{56B4E9}
\definecolor{oiBluishGreen}{HTML}{009E73}
\definecolor{oiYellow}{HTML}{D9C200}
\definecolor{oiBlue}{HTML}{0072B2}
\definecolor{oiVermilion}{HTML}{D55E00}
\definecolor{oiReddishPurple}{HTML}{CC79A7}

\pgfplotscreateplotcyclelist{okabeito-accessible}{
    {gray, dashed, mark=+, mark options={solid}, mark size=3.5pt, opacity=0.8}, 
    {oiOrange, dashed, mark=Mercedes star,mark options={solid}, mark size=4pt,opacity=0.8}, 
    {oiSkyBlue, dashed, mark=x, mark options={solid}, mark size=4pt,opacity=0.8}, 
    {oiBluishGreen, dashed, mark=star, mark options={solid}, mark size=4pt,opacity=0.6}, 
    {oiYellow, dashed, mark=asterisk, mark options={solid}, mark size=4.3pt,opacity=0.9}, 
    {oiBlue, dashed, mark=square, mark options={dashed}, mark size=3pt,opacity=0.6}, 
    {oiVermilion, dashed, mark=o, mark options={dashed}, mark size=3pt,opacity=0.8}, 
    {oiReddishPurple, dashed, mark=triangle, mark options={dashed},mark size=4pt,opacity=0.8} 
}

\pgfplotsset{
    cycle list name=okabeito-accessible,
    every axis plot/.append style={line width=4pt}
}

\setlist[itemize,1]{leftmargin=2.5ex}
\setlist[itemize,2]{leftmargin=2.5ex,label=$\circ$}

\newcommand{\ours}[0]{Cultivar}

\definecolor{annotationoriginal}{HTML}{f4d8e1}
\definecolor{annotationrephrased}{HTML}{d4e6d8}

\title{\textsc{\ours{}}: A Contrastive and Locale-Oriented Translation Benchmark \\ for Investigating Contamination and Localisation Robustness}

\author{
\vspace{1.6ex}
Pinzhen Chen\textsuperscript{1}\qquad
Koel Dutta Chowdhury\textsuperscript{2}\qquad
Xiaoya Xu\textsuperscript{1}\qquad
David Tan\textsuperscript{3}\\
\bfseries 
Doreen Osmelak\textsuperscript{3,4}\quad
Ona de Gibert\textsuperscript{5}\quad
Ariun-Erdene Tumurchuluun\quad
Ashok Urlana\textsuperscript{6,7}\\
\bfseries 
Fedor Sizov\textsuperscript{3}\quad
Hale Sirin\textsuperscript{8}\quad
Jesujoba O. Alabi\textsuperscript{3}\quad
Karrar Talib Abed\textsuperscript{9}\\
\bfseries 
Mateusz Klimaszewski\textsuperscript{10}\quad
Nikolay Bogoychev\textsuperscript{11}\quad
Niyati Bafna\textsuperscript{8}\quad
Patrícia Schmidtová\textsuperscript{12}\\
\bfseries
Preksha Manjunath Shanbhag\textsuperscript{1}\quad
Sherrie Shen\textsuperscript{13}\quad
Vilém Zouhar\textsuperscript{14}\\
\vspace{1.6ex}
\bfseries
Vivek Iyer\textsuperscript{13}\quad
Yasser Hamidullah\textsuperscript{15}\quad
Yusser Al Ghussin\textsuperscript{3}\qquad
Zheng Zhao\textsuperscript{13}\\
\textsuperscript{1}Queen's University Belfast,
\textsuperscript{2}University of Technology Nuremberg,
\textsuperscript{3}Saarland University\\
\textsuperscript{4}University of Melbourne,
\textsuperscript{5}University of Helsinki,
\textsuperscript{6}IIIT Hyderabad,
\textsuperscript{7}TCS Research\\
\textsuperscript{8}Johns Hopkins University,
\textsuperscript{9}Imam Ja'afar Al-Sadiq University, 
\textsuperscript{10}Cohere,
\textsuperscript{11}Last Token\\
\vspace{1.6ex}
\textsuperscript{12}Charles University,
\textsuperscript{13}University of Edinburgh,
\textsuperscript{14}ETH Zurich,
\textsuperscript{15}University of Zurich\\
\texttt{p.chen@qub.ac.uk}
}

\begin{document}
\maketitle

\begin{abstract}
Multilingual translation benchmarks are typically sourced in English and translated into other languages, treating language pairs as the unit of evaluation---a design that is prone to contamination over time and overlooks locale and cultural considerations. We therefore advocate for source-contrastive evaluation and instantiate it with \ours{}, a localised subset of FLORES, which enables locale-specific translation evaluation. When paired with unlocalised counterparts, performance discrepancy allows the probing of data contamination and localisation robustness. We benchmark 32 open-weight models and find that MT-specialised models are less robust, a few models potentially overfit FLORES, and models tend to translate US content better than that of other locales, regardless of language.\footnotenonumber{\hspace{-3ex}\twemoji{potted plant}\,\rainbow{\ours{}} is potted at \url{https://hf.co/datasets/pinzhenchen/Cultivar-flores}, harvestable under CC BY-SA 4.0.}
\end{abstract}

\section{Introduction}

Benchmarks are an essential element for developing translation systems. As machine translation technology reaches maturity and is deployed for more languages, recent years have seen a shift from English-centric evaluation, like the early WMT \citep{bojar-etal-2014-findings} and IWSLT \citep{cettolo-etal-2014-report} evaluation campaigns, towards multilingual test suites such as FLORES \citep{goyal-etal-2022-flores,nllb2024scaling}, NTREX \citep{federmann-etal-2022-ntrex}, and WMT24++ \citep{deutsch-etal-2025-wmt24}, enabling evaluation for hundreds of languages in arbitrary directions. 

Yet, massively multilingual test suites share a common limitation due to construction: regardless of language, each instance is anchored to a single information origin, which is typically English, and then translated. While this facilitates multiway parallelism, it introduces two vulnerabilities. First, because the original English content is publicly available, test sets become susceptible to data contamination over time \citep{magar-schwartz-2022-data,sainz-etal-2023-nlp,yao-etal-2024-data,tan-etal-2026-flores}. Second, 
using such test sets to evaluate translation into English creates a language-content mismatch \citep{chen-etal-2024-good-data}, which diverges from real-world use cases, where the input content is natively produced in the input language. BOUQuET \citep{andrews-etal-2025-bouquet} alleviates this problem by starting with original texts in eight distinct languages and then translating into more. A separate line of effort trades away multiway parallelism for naturally written, culturally grounded text, such as WMT since 2019 \citep{barrault-etal-2019-findings} and multilingual benchmarks like TyDi QA \citep{clark-etal-2020-tydi}, which improves the authenticity of evaluation from a multicultural perspective but loses comparability.

Nonetheless, we argue that source-origin content alone is insufficient. Existing test sets are fundamentally language-oriented, assuming that each language corresponds to a single cultural context. In practice, however, a language can often be used in multiple regions, with diaspora communities adding an extra layer of variation. Consequently, evaluating a model on a single test for each language either fails to capture other locales or receives an arbitrary, mixed-origin result. Thus, we argue that translation and multilingual evaluation should move beyond language-oriented tests towards locale-oriented tests, where multiple localisations of the same language are separately assessed to better reflect real-world use.

We therefore advocate a \emph{source-contrastive} paradigm that grounds an input in distinct contexts \citep{bawden-etal-2018-evaluating,futeral-etal-2023-tackling,alam-etal-2024-codet}. 
Instead of testing a single static sentence, an evaluation suite should feature paired instances that preserve sentence structure while systematically varying locale-specific content based on desired locations. Such a controlled design isolates a model's sensitivity to varied content from general translation capability. Crucially, it enables analyses that are impossible with conventional single-version tests: diagnosing benchmark contamination and quantifying true localisation robustness through paired performance discrepancies. To instantiate this paradigm, we present the \textit{\ours{}} benchmark, grounded in 27 locales, created through an LLM-human co-creation pipeline starting from FLORES. We benchmark 32 language and translation models and share our findings. To summarise, our contributions are:

\begin{itemize}[itemsep=0.3ex,topsep=0.3ex]

\item \textbf{Resource:} \ours{}, a multilingual translation test set grounded in locales, produced via a human-LLM co-creation pipeline. We preserve the comparability of massively multilingual test sets while incorporating native source content.

\item \textbf{Evaluation:} We evaluate 32 models on Cultivar using metrics like BLEU, chrF, and word recall. We report standalone locale-level scores and conduct a contrastive study with FLORES, quantifying localisation robustness and examining overfitting.

\item \textbf{Findings:} Performance of MT-specialised and smaller models tends to drop on localised content. Significant gaps between original and localised tests reveal potential FLORES overfitting. Two smaller-scale studies also show that models perform better on content grounded in the US than on localised content in the official language of a particular region.
\end{itemize}

\section{\ours{}}

We construct \ours{} through a three-stage pipeline that localises FLORES: 1) named entity-rich instance selection; 2) LLM-assisted paraphrasing; 3) human post-editing to ensure quality. Although we create \ours{} by localising FLORES, technically, the pipeline can be applied to any existing benchmark.

Unlike efforts that extend FLORES to new languages by translating from scratch \citep{dale-etal-2025-findings}, we formulate localisation as creative paraphrasing. Our design preserves the sentence structure of original instances while adapting cultural content to a specific location, thereby producing sentence pairs that remain directly comparable. 

An LLM-assisted annotation workflow is particularly suitable for this setting because localisation requires both creativity and diversity beyond a translation task, making creation from scratch difficult for humans. Our use of human post-editing ensures quality and cultural appropriateness. 

In this study, we only localise any-to-English translation directions. The resulting English target sentences across locales are no longer the same, but they remain comparable. Namely, it is still meaningful to compare error counts or even word-level metrics like BLEU between into-English directions.

\subsection{Named entity-rich instance identification}

Localisation primarily affects named entities such as people, organisations, locations, events, etc. We thus first identify named entities in each instance of the English FLORES dev set using spaCy \citep{honnibal2020spacy} and manually correct missing or erroneous annotations. This process finds 658 sentences (approximately 65\% of the dev set) containing at least one named entity. 

From these, we manually select 200 sentences having at least 2 named entities. We exclude instances dominated by numerals or topics such as medicine and science, where named entities are globally shared. The resulting named entity-rich subset forms the input to the LLM localisation step.

\subsection{LLM paraphrasing}
Given a source language of interest, for each selected English instance, an LLM generates a localised source-English pair from the original FLORES source-English pair and a target location. The model is instructed to preserve the sentence structure whenever possible while replacing content with appropriate alternatives for the new location. Our full prompt is provided in \Cref{fig:System and user prompts used for LLM localisation} in \Cref{sec:LLM Localisation Prompts}. We use GPT-5.5 (\texttt{gpt-5.5-2026-0423}).

\begin{figure*}[t]
    \centering\small
    \includegraphics[width=\linewidth]{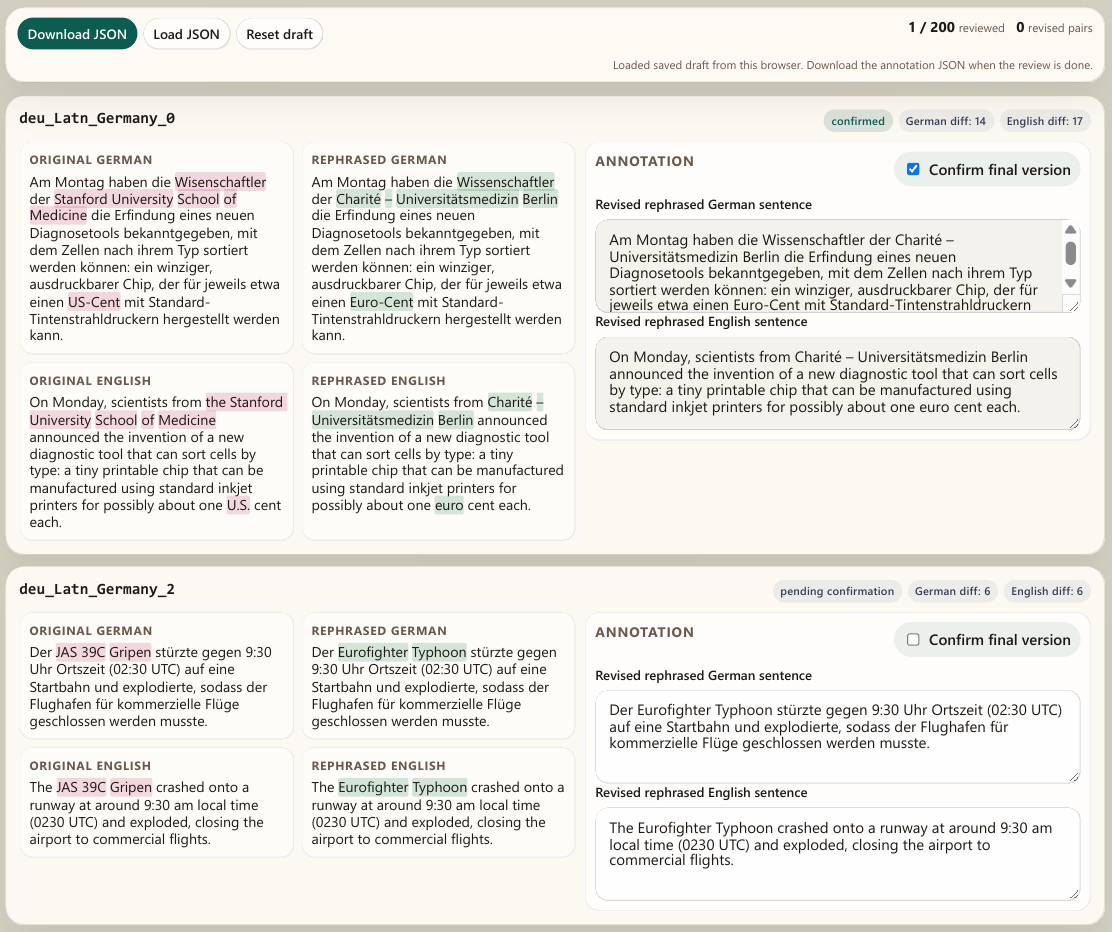}
    \caption{An illustration of the annotation interface and the content before and after localising German written in Latin script (\texttt{deu\_Latn}) to Germany, paired with English. The difference between the {\setlength{\fboxsep}{1pt}\colorbox{annotationoriginal}{original}} and {\setlength{\fboxsep}{1pt}\colorbox{annotationrephrased}{replaced}} content is highlighted. The annotation panel on the right is, by default, populated with the LLM-rephrased content, which the annotator can edit; as they edit, the highlights update accordingly. Once done, annotators check ``Confirm final version'' to freeze the edit box to finalise this sentence pair.}
    \label{fig:Annotation Interface}
\end{figure*}

\subsection{Human annotation}

\paragraph{Annotator profile} We use three criteria to select an annotator: 1) proficient in the source language; 2) proficient in the target language (English); and 3) they live(d) in the location. These requirements ensure both language competence and familiarity with local cultural conventions. 

The majority of annotators are NLP researchers with some understanding of NLP evaluation. All annotators are included as authors, as their contributions directly shape the benchmark and analyses.

\paragraph{Annotation process} We ask annotators to inspect each instance pair before and after LLM localisation and perform one of the three actions below:
\begin{enumerate}[noitemsep,topsep=0pt]
    \item accept the LLM paraphrase;
    \item post-edit the paraphrase;
    \item revert to the original FLORES entry.
\end{enumerate}
Similar to the LLM prompting stage, annotators are instructed to preserve the original sentence structure whenever possible to maintain comparability between the original and localised benchmarks. 
The first two options produce approved localised instances. Annotators were also allowed to use option 3 sparingly, when localisation is unnecessary, culturally inappropriate, or difficult.

Before annotation starts, annotators are instructed to read through a brief motivation and description of the project, annotation guidelines, and a few examples of what they do and do not need to edit. Detailed annotation instructions are presented in \Cref{fig:Project details and annotation instruction given to human annotators} in \Cref{sec:Instructions Presented to Annotators}. Annotation time ranges from 4 hours to 12 hours.

\begin{table*}[t!]
    \centering
    \small
    \setlength{\tabcolsep}{1.3ex}
    
\begin{tabular}{lllrrrrrrr}
\toprule
\multirow[b]{2}{*}{Lang}
& \multirow[b]{2}{*}{Script}
    & \multirow[b]{2}{*}{Location}
& \multirow[b]{2}{*}{\makecell{Kept LLM\\Paraphrase}}
& \multirow[b]{2}{*}{\makecell{Reverted to\\FLORES}}
& \multicolumn{3}{c}{Human Edited}
& \multicolumn{2}{c}{Average Edits} \\
\cmidrule(lr){6-8}\cmidrule(lr){9-10}
& & & &
& Source Only
& Target Only
& Both
& Source 
& English \\
\midrule
acm & Arab & Iraq & 176 & 0 & 14 & 1 & 9 & 4.80 & 4.28 \\
apc & Arab & Syria & 155 & 0 & 14 & 2 & 29 & 4.78 & 4.38 \\
ben & Beng & India & 146 & 0 & 41 & 0 & 13 & 7.67 & 4.19 \\
bul & Cyrl & Bulgaria & 161 & 0 & 2 & 0 & 37 & 4.41 & 4.14 \\
cat & Latn & Spain & 136 & 8 & 1 & 1 & 54 & 4.70 & 4.43 \\
ces & Latn & Czechia & 109 & 0 & 15 & 12 & 64 & 4.61 & 4.44 \\
cmn & Hans & China & 168 & 1 & 2 & 7 & 22 & 4.01 & 4.27 \\
 &  & Singapore & 153 & 1 & 7 & 5 & 34 & 4.47 & 4.72 \\
 &  & UK\textsuperscript{$\dagger$} & 170 & 7 & 10 & 0 & 13 & 4.37 & 4.22 \\
 &  & US\textsuperscript{$\dagger$} & 180 & 4 & 0 & 1 & 15 & 4.34 & 4.28 \\
deu & Latn & Germany & 117 & 0 & 3 & 2 & 78 & 4.48 & 4.46 \\
 &  & Switzerland & 154 & 1 & 6 & 1 & 38 & 4.66 & 4.50 \\
fil & Latn & Philippines & 193 & 1 & 4 & 0 & 2 & 4.24 & 4.15 \\
gom & Deva & India & 174 & 8 & 0 & 0 & 18 & 6.14 & 3.99 \\
hin & Deva & Germany\textsuperscript{$\dagger$} & 1 & 1 & 0 & 0 & 198 & 6.90 & 4.32 \\
 &  & India & 112 & 7 & 0 & 7 & 74 & 7.21 & 4.53 \\
 &  & US\textsuperscript{$\dagger$} & 113 & 34 & 8 & 0 & 45 & 7.44 & 4.47 \\
jpn & Jpan & Japan & 89 & 2 & 7 & 32 & 70 & 3.63 & 4.47 \\
khk & Cyrl & Mongolia & 68 & 0 & 53 & 14 & 65 & 5.25 & 4.46 \\
plt & Latn & Madagascar & 187 & 0 & 5 & 0 & 8 & 4.12 & 3.92 \\
pol & Latn & Poland & 176 & 3 & 0 & 19 & 2 & 4.28 & 4.12 \\
rus & Cyrl & Russia & 136 & 8 & 26 & 0 & 30 & 4.90 & 4.33 \\
slk & Latn & Slovakia & 144 & 1 & 3 & 1 & 51 & 4.68 & 4.29 \\
spa & Latn & Spain & 117 & 6 & 2 & 3 & 72 & 4.91 & 4.44 \\
tel & Telu & India & 173 & 0 & 5 & 2 & 20 & 7.30 & 4.36 \\
tur & Latn & Turkey & 147 & 6 & 13 & 1 & 33 & 4.21 & 4.19 \\
yor & Latn & Nigeria & 127 & 0 & 56 & 1 & 16 & 5.28 & 4.15 \\
\midrule
\multicolumn{3}{l}{Average} & 140.07 & 3.67 & 11.00 & 4.15 & 41.11 & 5.10 & 4.31 \\
\bottomrule
\end{tabular}
    \caption{Summary of human edits to LLM-localised FLORES instances: LLM paraphrases are kept as is, reverted to FLORES, or edited by the annotators. Localisation targets regions where a language holds official or de facto status, except for the four entries marked with a $\dagger$.}
    \label{tab:annotation-statistics}
\end{table*}

\paragraph{Annotation interface} Annotation is performed through a custom web interface shown as Figure~\ref{fig:Annotation Interface}. It displays the original and localised sentence pairs side by side while highlighting changed spans. The edit box is initialised with the LLM paraphrase, allowing annotators to efficiently post-edit the text. Differences and highlights are updated as edits are made.

After completing each entry, annotators finalise it by checking a confirmation box. This deliberate step acts as a check on attention and responsibility, as it encourages careful review and reduces the likelihood of inadvertent skipping.

\subsection{Annotated data statistics}

The final localised Cultivar benchmark consists of 27 language-script-location combinations---referred to as \textit{locales}. We represented them with the same language and writing script codes as FLORES, together with the country name where the data is grounded. There are 21 unique languages or dialects, 8 unique writing scripts, and 21 unique countries or regions. Each locale has 200 instances contributed by a single annotator. 

A breakdown of annotation statistics for each locale is in \Cref{tab:annotation-statistics}. On average, annotators accept 70\% of LLM-generated localisations without modification, post-edit roughly 28\%, and revert fewer than 2\% to the original FLORES entries. Consequently, over 98\% of \ours{} is newly localised content. A few outliers exist, for example: all but one of the LLM paraphrases are human-edited when Hindi is localised to Germany; only 7 and 13 are changed when Filipino and Malagasy are localised, respectively.

At the source-target pair level, we count a modified continuous string surrounded by original strings as an edit. We only count edits to an instance if the LLM paraphrase is revised---either reverted to FLORES or edited by the annotator. On average, there are 4 edits per English instance and 5 edits per source instance. However, the latter is likely inflated by the Indian languages in our dataset (\texttt{ben,gom,hin,tel}), all of which average around 7 edits per source. 

\section{Experimental Setup}

\subsection{Models}
We evaluate 32 open-weight models that can be deployed on no more than two NVIDIA A100 80GB GPUs. Our model selection considers four dimensions: 1) translation-specific or general-purpose; 2) 0.8--122B in size; 3) English-focused or multilingual; 4) made in or targeting diverse locations. We list them under three groups:
\begin{itemize}[itemsep=0.1ex,topsep=0pt]
\item Pre-trained machine translation models (MT): \texttt{nllb-200-distilled-\{600M,1.3B\}} \citep{nllb2024scaling} and \texttt{madlad400-\{3b,7b\}-mt} \citep{madlad400}.
\item LLM-based translation models (LLM+MT): \texttt{Seed-X-PPO-7B} \citep{cheng2025seed} and \texttt{Hy-MT2-\{1.8B,7B,30B-A3B\}} \citep{zheng2026hy}.
\item LLMs: \texttt{tiny-aya-global} \citep{salamanca2026tiny}, \texttt{aya-expanse-8b} \citep{dang2024aya}, \texttt{SmolLM-1.7B-Instruct} \citep{allal2024SmolLM}, \texttt{emma-500-llama3.1-8b-\{mono,bi\}} \citep{ji-etal-2026-data}, \texttt{Qwen3.5-\{0.8B,2B,4B,9B,27B,35B-A3B,122B-A10B-FP8\}} \citep{qwen3.5}, \texttt{Olmo-3-7B-Instruct} \citep{olmo2025olmo}, \texttt{gemma-3-\{1b,4b\}-it} \citep{gemma_2025}, \texttt{Llama-3.1-8B-instruct}, \texttt{Llama-3.2-\{1B,3B\}-instruct}, \texttt{Llama-3.3-70B-Instruct} \citep{grattafiori2024llama}, \texttt{Phi-4-mini-instruct} \citep{abouelenin2025phi}, \texttt{Ministral-3-\{3B,8B\}-Instruct-2512} \citep{liu2026ministral}, and \texttt{EuroLLM-\{1.7B-Instruct,\\9B-instruct-2512\}} \citep{martins2025eurollm}.
\end{itemize}

\subsection{Evaluation}\label{subsec:metrics}
\paragraph{BLEU and chrF} We compute BLEU \citep{papineni-etal-2002-bleu} and chrF \citep{popovic-2015-chrf}, two standard automatic metrics for machine translation evaluation. Specifically, we use spBLEU with \texttt{flores200} tokenizer and chrF with 0 word order, as implemented in sacrebleu \citep{post-2018-call}. This way, the metrics operate on subwords and characters, respectively, without complications due to tokenization for certain languages.

\paragraph{$\Delta_\text{BLEU}$ and $\Delta_\text{chrF}$ as measures of robustness} 
Our primary research objective is not to discover the best translation models, but to quantify their robustness via results before and after localisation. Therefore, we define a performance discrepancy between the localised and original tests as:
\begin{align}
\nonumber\Delta_\text{BLEU} &= \text{BLEU}_{\text{localised}} -\text{BLEU}_{\text{original}}\\
\nonumber\Delta_\text{chrF} &= \text{chrF}_{\text{localised}} - \text{chrF}_{\text{original}}
\end{align}
As $\Delta$ becomes increasingly negative, the model exhibits greater performance degradation when translating localised source texts. Importantly, $\Delta$ is largely invariant to translation quality---a model that performs poorly or strongly on both test sets will receive a $\Delta\to0$ for its consistent behaviour.

\paragraph{$\overline{\Delta}_\text{BLEU}$ and $\overline{\Delta}_\text{chrF}$} To aggregate over multiple locales after localisation, we pair all locales with their language, forming pairs $i=1,2,\cdots,n$ and compute a mean pairwise $\Delta_\text{metric}$ applicable to both BLEU and chrF:
\begin{equation}
\overline{\Delta}_\text{metric}=\frac{1}{n}\sum^{n}_{i=1}(\text{score}_{\text{localised},i} - \text{score}_{\text{original},i})\nonumber
\end{equation}
Likewise, the mean $\Delta_\text{metric}$ can be averaged across models but without requiring pairing.

\paragraph{Locale-specific word recall} Standard metrics reveal overall quality (difference) but provide little focus on locale-specific content. Since both LLMs and human annotators were instructed to avoid changing the sentence structure, most word edits would be directly related to the effort of language-location grounding. We therefore introduce a simple lexical analysis that looks at the translation of a localised input, its original FLORES reference, and the localised \ours{} reference: 
\begin{itemize}[itemsep=0.3ex,topsep=0.3ex]
    \item \textbf{False positives (FP):} unique terms in the original FLORES reference that appear in the translation of a localised instance. These should no longer appear after localisation. 
    \item \textbf{False negatives (FN)}: unique terms in the localised \ours{} reference that do not appear in the translation. Since the localisation process introduced such words, they should appear. 
\end{itemize}
A high false positive indicates that a model continues to generate terms in the original FLORES despite their removal during localisation, suggesting memorization.  Conversely, a high false negative means that models fail to generate newly introduced locale-specific content. Nonetheless, as multiple valid translations may exist for a localised concept, false negatives should be interpreted as an upper bound on localisation failures rather than an exact error rate.

\begin{table}[t!]
    \centering\small
    \setlength{\tabcolsep}{1.4ex}
    \begin{tabular}{llcccc}
\toprule
\multirow[b]{2}{*}{Original} & \multirow[b]{2}{*}{Localised} & \multicolumn{2}{c}{BLEU} & \multicolumn{2}{c}{chrF} \\[-1pt]
\cmidrule(lr){3-4}
\cmidrule(lr){5-6}
 &  & $r_{s}$ & $\tau_b$ & $r_{s}$ & $\tau_b$ \\
\midrule
acm\_Arab & Iraq & \cellcolor{Gold2!15!SeaGreen3!85}0.97 & \cellcolor{Gold2!38!SeaGreen3!62}0.89 & \cellcolor{Gold2!13!SeaGreen3!87}0.97 & \cellcolor{Gold2!34!SeaGreen3!66}0.90 \\
apc\_Arab & Syria & \cellcolor{Gold2!15!SeaGreen3!85}0.97 & \cellcolor{Gold2!39!SeaGreen3!61}0.88 & \cellcolor{Gold2!11!SeaGreen3!89}0.98 & \cellcolor{Gold2!30!SeaGreen3!70}0.91 \\
ben\_Beng & India & \cellcolor{Gold2!10!SeaGreen3!90}0.98 & \cellcolor{Gold2!26!SeaGreen3!74}0.92 & \cellcolor{Gold2!10!SeaGreen3!90}0.98 & \cellcolor{Gold2!26!SeaGreen3!74}0.92 \\
bul\_Cyrl & Bulgaria & \cellcolor{Gold2!33!SeaGreen3!67}0.93 & \cellcolor{Gold2!67!SeaGreen3!33}0.80 & \cellcolor{Gold2!30!SeaGreen3!70}0.94 & \cellcolor{Gold2!56!SeaGreen3!44}0.83 \\
cat\_Latn & Spain & \cellcolor{Gold2!16!SeaGreen3!84}0.97 & \cellcolor{Gold2!42!SeaGreen3!58}0.88 & \cellcolor{Gold2!11!SeaGreen3!89}0.98 & \cellcolor{Gold2!31!SeaGreen3!69}0.91 \\
ces\_Latn & Czechia & \cellcolor{Gold2!12!SeaGreen3!88}0.98 & \cellcolor{Gold2!38!SeaGreen3!62}0.89 & \cellcolor{Gold2!10!SeaGreen3!90}0.98 & \cellcolor{Gold2!32!SeaGreen3!68}0.90 \\
cmn\_Hans & China & \cellcolor{Gold2!23!SeaGreen3!77}0.95 & \cellcolor{Gold2!54!SeaGreen3!46}0.84 & \cellcolor{Gold2!18!SeaGreen3!82}0.96 & \cellcolor{Gold2!50!SeaGreen3!50}0.85 \\
cmn\_Hans & Singapore & \cellcolor{Gold2!28!SeaGreen3!72}0.94 & \cellcolor{Gold2!66!SeaGreen3!34}0.80 & \cellcolor{Gold2!25!SeaGreen3!75}0.95& \cellcolor{Gold2!52!SeaGreen3!48}0.84 \\
cmn\_Hans & UK & \cellcolor{Gold2!24!SeaGreen3!76}0.95 & \cellcolor{Gold2!51!SeaGreen3!49}0.85 & \cellcolor{Gold2!12!SeaGreen3!88}0.98 & \cellcolor{Gold2!34!SeaGreen3!66}0.90 \\
cmn\_Hans & US & \cellcolor{Gold2!8!SeaGreen3!92}0.98 & \cellcolor{Gold2!30!SeaGreen3!70}0.91 & \cellcolor{Gold2!7!SeaGreen3!93}0.99 & \cellcolor{Gold2!27!SeaGreen3!73}0.92 \\
deu\_Latn & Germany & \cellcolor{Gold2!10!SeaGreen3!90}0.98 & \cellcolor{Gold2!34!SeaGreen3!66}0.90 & \cellcolor{Gold2!6!SeaGreen3!94}0.99 & \cellcolor{Gold2!24!SeaGreen3!76}0.93 \\
deu\_Latn & Switzerland & \cellcolor{Gold2!16!SeaGreen3!84}0.97 & \cellcolor{Gold2!42!SeaGreen3!58}0.88 & \cellcolor{Gold2!11!SeaGreen3!89}0.98 & \cellcolor{Gold2!34!SeaGreen3!66}0.90 \\
fil\_Latn & Philippines & \cellcolor{Gold2!14!SeaGreen3!86}0.97 & \cellcolor{Gold2!27!SeaGreen3!73}0.92 & \cellcolor{Gold2!11!SeaGreen3!89}0.98 & \cellcolor{Gold2!23!SeaGreen3!77}0.93 \\
gom\_Deva & India & \cellcolor{Gold2!23!SeaGreen3!77}0.95 & \cellcolor{Gold2!50!SeaGreen3!50}0.85 & \cellcolor{Gold2!16!SeaGreen3!84}0.97 & \cellcolor{Gold2!36!SeaGreen3!64}0.89 \\
hin\_Deva & Germany & \cellcolor{Gold2!22!SeaGreen3!78}0.96 & \cellcolor{Gold2!44!SeaGreen3!56}0.87 & \cellcolor{Gold2!17!SeaGreen3!83}0.97 &\cellcolor{Gold2!36!SeaGreen3!64}0.89 \\
hin\_Deva & India & \cellcolor{Gold2!24!SeaGreen3!76}0.95 & \cellcolor{Gold2!48!SeaGreen3!52}0.85 & \cellcolor{Gold2!18!SeaGreen3!82}0.96 & \cellcolor{Gold2!40!SeaGreen3!60}0.88 \\
hin\_Deva & US & \cellcolor{Gold2!12!SeaGreen3!88}0.98 & \cellcolor{Gold2!27!SeaGreen3!73}0.92 & \cellcolor{Gold2!15!SeaGreen3!85}0.97 & \cellcolor{Gold2!32!SeaGreen3!68}0.90 \\
jpn\_Jpan & Japan & \cellcolor{Gold2!30!SeaGreen3!70}0.94 & \cellcolor{Gold2!60!SeaGreen3!40}0.82 & \cellcolor{Gold2!25!SeaGreen3!75}0.95 & \cellcolor{Gold2!55!SeaGreen3!45}0.83 \\
khk\_Cyrl & Mongolia & \cellcolor{Gold2!16!SeaGreen3!84}0.97 & \cellcolor{Gold2!46!SeaGreen3!54}0.86 & \cellcolor{Gold2!12!SeaGreen3!88}0.98 & \cellcolor{Gold2!35!SeaGreen3!65}0.90 \\
plt\_Latn & Madagascar & \cellcolor{Gold2!8!SeaGreen3!92}0.98 & \cellcolor{Gold2!27!SeaGreen3!73}0.92 & \cellcolor{Gold2!4!SeaGreen3!96}0.99 & \cellcolor{Gold2!22!SeaGreen3!78}0.94 \\
pol\_Latn & Poland & \cellcolor{Gold2!33!SeaGreen3!67}0.93 & \cellcolor{Gold2!59!SeaGreen3!41}0.82 & \cellcolor{Gold2!19!SeaGreen3!81}0.96 & \cellcolor{Gold2!48!SeaGreen3!52}0.85 \\
rus\_Cyrl & Russia & \cellcolor{Gold2!18!SeaGreen3!82}0.96 & \cellcolor{Gold2!44!SeaGreen3!56}0.87 & \cellcolor{Gold2!13!SeaGreen3!87}0.97 & \cellcolor{Gold2!42!SeaGreen3!58}0.88 \\
slk\_Latn & Slovakia & \cellcolor{Gold2!15!SeaGreen3!85}0.97 & \cellcolor{Gold2!38!SeaGreen3!62}0.89 & \cellcolor{Gold2!13!SeaGreen3!87}0.97 & \cellcolor{Gold2!36!SeaGreen3!64}0.89 \\
spa\_Latn & Spain & \cellcolor{Gold2!15!SeaGreen3!85}0.97 & \cellcolor{Gold2!42!SeaGreen3!58}0.88 & \cellcolor{Gold2!12!SeaGreen3!88}0.98 & \cellcolor{Gold2!31!SeaGreen3!69}0.91 \\
tel\_Telu & India & \cellcolor{Gold2!19!SeaGreen3!81}0.96 & \cellcolor{Gold2!35!SeaGreen3!65}0.90 & \cellcolor{Gold2!15!SeaGreen3!85}0.97 & \cellcolor{Gold2!35!SeaGreen3!65}0.90 \\
tur\_Latn & Turkey & \cellcolor{Gold2!17!SeaGreen3!83}0.97 & \cellcolor{Gold2!43!SeaGreen3!57}0.87 & \cellcolor{Gold2!10!SeaGreen3!90}0.98 & \cellcolor{Gold2!32!SeaGreen3!68}0.90 \\
yor\_Latn & Nigeria & \cellcolor{Gold2!6!SeaGreen3!94}0.99 & \cellcolor{Gold2!27!SeaGreen3!73}0.92 & \cellcolor{Gold2!3!SeaGreen3!97}0.99 & \cellcolor{Gold2!15!SeaGreen3!85}0.96 \\
\midrule
\multicolumn{2}{l}{Average} & \cellcolor{Gold2!9!SeaGreen3!91}0.98 & \cellcolor{Gold2!31!SeaGreen3!69}0.91 & \cellcolor{Gold2!12!SeaGreen3!88}0.98 & \cellcolor{Gold2!36!SeaGreen3!64}0.89 \\
\bottomrule
\end{tabular}
    
    \caption{Spearman's $r_s$ and Kendall's $\tau_b$ for BLEU and chrF, measuring the alignment between rankings of 32 models based on the original FLORES and the localised \ours{} when translating into English.}
    \label{tab:ranking correlations}
\end{table}

\begin{table*}[t!]
\centering\small
\setlength{\tabcolsep}{1.35ex}
\begin{tabular}{llcccccccc}
\toprule
 \multicolumn{1}{c}{\multirow[b]{2}{*}{\textbf{Type}}} & \multicolumn{1}{c}{\multirow[b]{2}{*}{\textbf{Model}}} & \multicolumn{2}{c}{\textbf{Pairwise}} & \multicolumn{2}{c}{\textbf{FLORES}} & \multicolumn{4}{c}{\textbf{\ours{}}}    
 \\[-1pt]
\cmidrule(lr){3-4}\cmidrule(lr){5-6}\cmidrule(lr){7-10}
 & & \textbf{$\overline{\Delta}_\text{BLEU}\!\downarrow$} & \textbf{$\overline{\Delta}_\text{chrF}$} & \textbf{BLEU} & \textbf{chrF} & \textbf{BLEU} & \textbf{chrF} & \textbf{F Pos} & \textbf{F Neg}\\
\midrule
LLM & Qwen3.5-35B-A3B & \cellcolor{Gold2!0!SeaGreen3!100} \phantom{-}1.70 & \cellcolor{Gold2!2!SeaGreen3!98} \phantom{-}0.20  & \cellcolor{Gold2!35!SeaGreen3!65} 41.46 & \cellcolor{Gold2!27!SeaGreen3!73} 64.31  & \cellcolor{Gold2!9!SeaGreen3!91} 43.34 & \cellcolor{Gold2!10!SeaGreen3!90} 64.86  & \cellcolor{Gold2!5!SeaGreen3!95} 0.0069 & \cellcolor{Gold2!4!SeaGreen3!96} 0.1186 \\
LLM & Llama-3.1-8B-instruct & \cellcolor{Gold2!2!SeaGreen3!98} \phantom{-}1.42 & \cellcolor{Gold2!0!SeaGreen3!100} \phantom{-}0.31  & \cellcolor{Gold2!53!SeaGreen3!47} 35.36 & \cellcolor{Gold2!45!SeaGreen3!55} 59.39  & \cellcolor{Gold2!31!SeaGreen3!69} 37.58 & \cellcolor{Gold2!28!SeaGreen3!72} 60.30  & \cellcolor{Gold2!18!SeaGreen3!82} 0.0076 & \cellcolor{Gold2!20!SeaGreen3!80} 0.1717 \\
LLM & Ministral-3-8B-Instruct-2512 & \cellcolor{Gold2!3!SeaGreen3!97} \phantom{-}1.40 & \cellcolor{Gold2!5!SeaGreen3!95} -0.04  & \cellcolor{Gold2!54!SeaGreen3!46} 35.25 & \cellcolor{Gold2!42!SeaGreen3!58} 60.10  & \cellcolor{Gold2!32!SeaGreen3!68} 37.40 & \cellcolor{Gold2!26!SeaGreen3!74} 60.75  & \cellcolor{Gold2!37!SeaGreen3!63} 0.0087 & \cellcolor{Gold2!19!SeaGreen3!81} 0.1690 \\
LLM & EuroLLM-9B-instruct-2512 & \cellcolor{Gold2!3!SeaGreen3!97} \phantom{-}1.36 & \cellcolor{Gold2!5!SeaGreen3!95} -0.03  & \cellcolor{Gold2!48!SeaGreen3!52} 36.98 & \cellcolor{Gold2!45!SeaGreen3!55} 59.38  & \cellcolor{Gold2!24!SeaGreen3!76} 39.61 & \cellcolor{Gold2!27!SeaGreen3!73} 60.63  & \cellcolor{Gold2!20!SeaGreen3!80} 0.0077 & \cellcolor{Gold2!21!SeaGreen3!79} 0.1753 \\
LLM & Phi-4-mini-instruct & \cellcolor{Gold2!6!SeaGreen3!94} \phantom{-}1.07 & \cellcolor{Gold2!13!SeaGreen3!87} -0.53  & \cellcolor{Gold2!71!SeaGreen3!29} 29.25 & \cellcolor{Gold2!60!SeaGreen3!40} 54.96  & \cellcolor{Gold2!54!SeaGreen3!46} 31.57 & \cellcolor{Gold2!47!SeaGreen3!53} 55.86  & \cellcolor{Gold2!46!SeaGreen3!54} 0.0092 & \cellcolor{Gold2!46!SeaGreen3!54} 0.2579 \\
LLM & gemma-3-4b-it & \cellcolor{Gold2!6!SeaGreen3!94} \phantom{-}0.96 & \cellcolor{Gold2!8!SeaGreen3!92} -0.24  & \cellcolor{Gold2!51!SeaGreen3!49} 35.95 & \cellcolor{Gold2!41!SeaGreen3!59} 60.47  & \cellcolor{Gold2!31!SeaGreen3!69} 37.63 & \cellcolor{Gold2!25!SeaGreen3!75} 61.01  & \cellcolor{Gold2!25!SeaGreen3!75} 0.0080 & \cellcolor{Gold2!26!SeaGreen3!74} 0.1908 \\
LLM & Qwen3.5-122B-A10B-FP8 & \cellcolor{Gold2!7!SeaGreen3!93} \phantom{-}0.92 & \cellcolor{Gold2!7!SeaGreen3!93} -0.17  & \cellcolor{Gold2!30!SeaGreen3!70} 43.26 & \cellcolor{Gold2!22!SeaGreen3!78} 65.71  & \cellcolor{Gold2!5!SeaGreen3!95} 44.38 & \cellcolor{Gold2!6!SeaGreen3!94} 65.80  & \cellcolor{Gold2!2!SeaGreen3!98} 0.0067 & \cellcolor{Gold2!0!SeaGreen3!100} 0.1039 \\
LLM & Llama-3.2-3B-instruct & \cellcolor{Gold2!7!SeaGreen3!93} \phantom{-}0.89 & \cellcolor{Gold2!10!SeaGreen3!90} -0.35  & \cellcolor{Gold2!63!SeaGreen3!37} 32.22 & \cellcolor{Gold2!57!SeaGreen3!43} 56.00  & \cellcolor{Gold2!44!SeaGreen3!56} 34.33 & \cellcolor{Gold2!42!SeaGreen3!58} 56.85  & \cellcolor{Gold2!50!SeaGreen3!50} 0.0094 & \cellcolor{Gold2!40!SeaGreen3!60} 0.2394 \\
LLM & Qwen3.5-27B & \cellcolor{Gold2!8!SeaGreen3!92} \phantom{-}0.80 & \cellcolor{Gold2!9!SeaGreen3!91} -0.25  & \cellcolor{Gold2!31!SeaGreen3!69} 42.70 & \cellcolor{Gold2!24!SeaGreen3!76} 65.14  & \cellcolor{Gold2!8!SeaGreen3!92} 43.79 & \cellcolor{Gold2!8!SeaGreen3!92} 65.25  & \cellcolor{Gold2!5!SeaGreen3!95} 0.0069 & \cellcolor{Gold2!3!SeaGreen3!97} 0.1149 \\
LLM & Olmo-3-7B-Instruct & \cellcolor{Gold2!10!SeaGreen3!90} \phantom{-}0.61 & \cellcolor{Gold2!12!SeaGreen3!88} -0.48  & \cellcolor{Gold2!78!SeaGreen3!22} 27.05 & \cellcolor{Gold2!66!SeaGreen3!34} 53.42  & \cellcolor{Gold2!65!SeaGreen3!35} 28.75 & \cellcolor{Gold2!54!SeaGreen3!46} 54.11  & \cellcolor{Gold2!86!SeaGreen3!14} 0.0114 & \cellcolor{Gold2!64!SeaGreen3!36} 0.3205 \\
LLM & Qwen3.5-9B & \cellcolor{Gold2!10!SeaGreen3!90} \phantom{-}0.61 & \cellcolor{Gold2!10!SeaGreen3!90} -0.37  & \cellcolor{Gold2!41!SeaGreen3!59} 39.29 & \cellcolor{Gold2!34!SeaGreen3!66} 62.51  & \cellcolor{Gold2!21!SeaGreen3!79} 40.38 & \cellcolor{Gold2!19!SeaGreen3!81} 62.68  & \cellcolor{Gold2!20!SeaGreen3!80} 0.0077 & \cellcolor{Gold2!14!SeaGreen3!86} 0.1520 \\
LLM & Qwen3.5-2B & \cellcolor{Gold2!10!SeaGreen3!90} \phantom{-}0.56 & \cellcolor{Gold2!11!SeaGreen3!89} -0.40  & \cellcolor{Gold2!77!SeaGreen3!23} 27.46 & \cellcolor{Gold2!69!SeaGreen3!31} 52.46  & \cellcolor{Gold2!66!SeaGreen3!34} 28.58 & \cellcolor{Gold2!60!SeaGreen3!40} 52.65  & \cellcolor{Gold2!57!SeaGreen3!43} 0.0098 & \cellcolor{Gold2!56!SeaGreen3!44} 0.2928 \\
LLM & Ministral-3-3B-Instruct-2512 & \cellcolor{Gold2!10!SeaGreen3!90} \phantom{-}0.52 & \cellcolor{Gold2!10!SeaGreen3!90} -0.38  & \cellcolor{Gold2!70!SeaGreen3!30} 29.72 & \cellcolor{Gold2!58!SeaGreen3!42} 55.71  & \cellcolor{Gold2!56!SeaGreen3!44} 31.07 & \cellcolor{Gold2!45!SeaGreen3!55} 56.20  & \cellcolor{Gold2!61!SeaGreen3!39} 0.0100 & \cellcolor{Gold2!37!SeaGreen3!63} 0.2291 \\
LLM & EuroLLM-1.7B-Instruct & \cellcolor{Gold2!11!SeaGreen3!89} \phantom{-}0.48 & \cellcolor{Gold2!11!SeaGreen3!89} -0.43  & \cellcolor{Gold2!76!SeaGreen3!24} 27.83 & \cellcolor{Gold2!78!SeaGreen3!22} 49.99  & \cellcolor{Gold2!59!SeaGreen3!41} 30.45 & \cellcolor{Gold2!62!SeaGreen3!38} 51.99  & \cellcolor{Gold2!29!SeaGreen3!71} 0.0082 & \cellcolor{Gold2!58!SeaGreen3!42} 0.2978 \\
LLM & Qwen3.5-4B & \cellcolor{Gold2!11!SeaGreen3!89} \phantom{-}0.42 & \cellcolor{Gold2!14!SeaGreen3!86} -0.64  & \cellcolor{Gold2!53!SeaGreen3!47} 35.33 & \cellcolor{Gold2!43!SeaGreen3!57} 59.82  & \cellcolor{Gold2!36!SeaGreen3!64} 36.32 & \cellcolor{Gold2!30!SeaGreen3!70} 59.78  & \cellcolor{Gold2!29!SeaGreen3!71} 0.0082 & \cellcolor{Gold2!26!SeaGreen3!74} 0.1916 \\
LLM & Llama-3.2-1B-instruct & \cellcolor{Gold2!12!SeaGreen3!88} \phantom{-}0.30 & \cellcolor{Gold2!19!SeaGreen3!81} -0.93  & \cellcolor{Gold2!94!SeaGreen3!6} 21.82 & \cellcolor{Gold2!96!SeaGreen3!4} 44.89  & \cellcolor{Gold2!86!SeaGreen3!14} 23.21 & \cellcolor{Gold2!90!SeaGreen3!10} 45.32  & \cellcolor{Gold2!57!SeaGreen3!43} 0.0098 & \cellcolor{Gold2!100!SeaGreen3!0} 0.4400 \\
LLM & gemma-3-1b-it & \cellcolor{Gold2!13!SeaGreen3!87} \phantom{-}0.27 & \cellcolor{Gold2!17!SeaGreen3!83} -0.79  & \cellcolor{Gold2!83!SeaGreen3!17} 25.35 & \cellcolor{Gold2!75!SeaGreen3!25} 50.84  & \cellcolor{Gold2!73!SeaGreen3!27} 26.63 & \cellcolor{Gold2!66!SeaGreen3!34} 51.19  & \cellcolor{Gold2!55!SeaGreen3!45} 0.0097 & \cellcolor{Gold2!78!SeaGreen3!22} 0.3675 \\
LLM & tiny-aya-global & \cellcolor{Gold2!15!SeaGreen3!85} \phantom{-}0.03 & \cellcolor{Gold2!18!SeaGreen3!82} -0.88  & \cellcolor{Gold2!61!SeaGreen3!39} 32.90 & \cellcolor{Gold2!54!SeaGreen3!46} 56.88  & \cellcolor{Gold2!47!SeaGreen3!53} 33.37 & \cellcolor{Gold2!44!SeaGreen3!56} 56.52  & \cellcolor{Gold2!71!SeaGreen3!29} 0.0106 & \cellcolor{Gold2!52!SeaGreen3!48} 0.2780 \\
LLM & SmolLM-1.7B-Instruct & \cellcolor{Gold2!15!SeaGreen3!85} -0.02 & \cellcolor{Gold2!4!SeaGreen3!96} \phantom{-}0.07  & \cellcolor{Gold2!100!SeaGreen3!0} \phantom{0}3.92 & \cellcolor{Gold2!100!SeaGreen3!0} 20.15  & \cellcolor{Gold2!100!SeaGreen3!0} \phantom{0}4.38 & \cellcolor{Gold2!100!SeaGreen3!0} 21.83  & \cellcolor{Gold2!34!SeaGreen3!66} 0.0085 & \cellcolor{Gold2!100!SeaGreen3!0} 0.6930 \\
LLM & emma-500-llama3.1-8b-mono & \cellcolor{Gold2!15!SeaGreen3!85} -0.03 & \cellcolor{Gold2!19!SeaGreen3!81} -0.96  & \cellcolor{Gold2!50!SeaGreen3!50} 36.58 & \cellcolor{Gold2!44!SeaGreen3!56} 59.64  & \cellcolor{Gold2!32!SeaGreen3!68} 37.32 & \cellcolor{Gold2!33!SeaGreen3!67} 59.25  & \cellcolor{Gold2!46!SeaGreen3!54} 0.0092 & \cellcolor{Gold2!33!SeaGreen3!67} 0.2155 \\
LLM & emma-500-llama3.1-8b-bi & \cellcolor{Gold2!16!SeaGreen3!84} -0.10 & \cellcolor{Gold2!16!SeaGreen3!84} -0.75  & \cellcolor{Gold2!30!SeaGreen3!70} 43.20 & \cellcolor{Gold2!24!SeaGreen3!76} 65.18  & \cellcolor{Gold2!8!SeaGreen3!92} 43.57 & \cellcolor{Gold2!10!SeaGreen3!90} 64.83  & \cellcolor{Gold2!14!SeaGreen3!86} 0.0074 & \cellcolor{Gold2!16!SeaGreen3!84} 0.1569 \\
MT & madlad400-7b-mt & \cellcolor{Gold2!20!SeaGreen3!80} -0.53 & \cellcolor{Gold2!25!SeaGreen3!75} -1.34  & \cellcolor{Gold2!32!SeaGreen3!68} 42.38 & \cellcolor{Gold2!27!SeaGreen3!73} 64.31  & \cellcolor{Gold2!12!SeaGreen3!88} 42.67 & \cellcolor{Gold2!14!SeaGreen3!86} 63.78  & \cellcolor{Gold2!41!SeaGreen3!59} 0.0089 & \cellcolor{Gold2!29!SeaGreen3!71} 0.2001 \\
LLM & Llama-3.3-70B-Instruct & \cellcolor{Gold2!20!SeaGreen3!80} -0.58 & \cellcolor{Gold2!20!SeaGreen3!80} -0.98  & \cellcolor{Gold2!27!SeaGreen3!73} 44.16 & \cellcolor{Gold2!22!SeaGreen3!78} 65.84  & \cellcolor{Gold2!5!SeaGreen3!95} 44.40 & \cellcolor{Gold2!7!SeaGreen3!93} 65.49  & \cellcolor{Gold2!12!SeaGreen3!88} 0.0073 & \cellcolor{Gold2!5!SeaGreen3!95} 0.1214 \\
LLM & Qwen3.5-0.8B & \cellcolor{Gold2!21!SeaGreen3!79} -0.70 & \cellcolor{Gold2!26!SeaGreen3!74} -1.41  & \cellcolor{Gold2!100!SeaGreen3!0} 19.70 & \cellcolor{Gold2!100!SeaGreen3!0} 43.80  & \cellcolor{Gold2!100!SeaGreen3!0} 19.66 & \cellcolor{Gold2!100!SeaGreen3!0} 42.83  & \cellcolor{Gold2!71!SeaGreen3!29} 0.0106 & \cellcolor{Gold2!100!SeaGreen3!0} 0.4403 \\
MT & nllb-200-distilled-1.3B & \cellcolor{Gold2!22!SeaGreen3!78} -0.79 & \cellcolor{Gold2!26!SeaGreen3!74} -1.43  & \cellcolor{Gold2!35!SeaGreen3!65} 41.44 & \cellcolor{Gold2!29!SeaGreen3!71} 63.76  & \cellcolor{Gold2!19!SeaGreen3!81} 40.71 & \cellcolor{Gold2!18!SeaGreen3!82} 62.75  & \cellcolor{Gold2!50!SeaGreen3!50} 0.0094 & \cellcolor{Gold2!40!SeaGreen3!60} 0.2382 \\
MT & madlad400-3b-mt & \cellcolor{Gold2!23!SeaGreen3!77} -0.90 & \cellcolor{Gold2!27!SeaGreen3!73} -1.50  & \cellcolor{Gold2!37!SeaGreen3!63} 40.88 & \cellcolor{Gold2!31!SeaGreen3!69} 63.25  & \cellcolor{Gold2!19!SeaGreen3!81} 40.84 & \cellcolor{Gold2!19!SeaGreen3!81} 62.52  & \cellcolor{Gold2!46!SeaGreen3!54} 0.0092 & \cellcolor{Gold2!36!SeaGreen3!64} 0.2266 \\
MT & nllb-200-distilled-600M & \cellcolor{Gold2!26!SeaGreen3!74} -1.21 & \cellcolor{Gold2!36!SeaGreen3!64} -2.05  & \cellcolor{Gold2!47!SeaGreen3!53} 37.40 & \cellcolor{Gold2!39!SeaGreen3!61} 60.95  & \cellcolor{Gold2!35!SeaGreen3!65} 36.60 & \cellcolor{Gold2!32!SeaGreen3!68} 59.51  & \cellcolor{Gold2!57!SeaGreen3!43} 0.0098 & \cellcolor{Gold2!56!SeaGreen3!44} 0.2924 \\
LLM+MT & Hy-MT2-1.8B & \cellcolor{Gold2!33!SeaGreen3!67} -2.06 & \cellcolor{Gold2!36!SeaGreen3!64} -2.09  & \cellcolor{Gold2!44!SeaGreen3!56} 38.54 & \cellcolor{Gold2!37!SeaGreen3!63} 61.45  & \cellcolor{Gold2!31!SeaGreen3!69} 37.77 & \cellcolor{Gold2!27!SeaGreen3!73} 60.68  & \cellcolor{Gold2!57!SeaGreen3!43} 0.0098 & \cellcolor{Gold2!36!SeaGreen3!64} 0.2243 \\
LLM+MT & Hy-MT2-7B & \cellcolor{Gold2!34!SeaGreen3!66} -2.16 & \cellcolor{Gold2!36!SeaGreen3!64} -2.06  & \cellcolor{Gold2!27!SeaGreen3!73} 44.29 & \cellcolor{Gold2!23!SeaGreen3!77} 65.54  & \cellcolor{Gold2!11!SeaGreen3!89} 42.98 & \cellcolor{Gold2!12!SeaGreen3!88} 64.38  & \cellcolor{Gold2!46!SeaGreen3!54} 0.0092 & \cellcolor{Gold2!18!SeaGreen3!82} 0.1661 \\
LLM+MT & Hy-MT2-30B-A3B & \cellcolor{Gold2!35!SeaGreen3!65} -2.28 & \cellcolor{Gold2!38!SeaGreen3!62} -2.21  & \cellcolor{Gold2!17!SeaGreen3!83} 47.46 & \cellcolor{Gold2!14!SeaGreen3!86} 68.17  & \cellcolor{Gold2!0!SeaGreen3!100} 45.77 & \cellcolor{Gold2!3!SeaGreen3!97} 66.58  & \cellcolor{Gold2!0!SeaGreen3!100} 0.0066 & \cellcolor{Gold2!5!SeaGreen3!95} 0.1217 \\
LLM & aya-expanse-8b & \cellcolor{Gold2!93!SeaGreen3!7} -8.87 & \cellcolor{Gold2!93!SeaGreen3!7} -5.84  & \cellcolor{Gold2!8!SeaGreen3!92} 50.51 & \cellcolor{Gold2!16!SeaGreen3!84} 67.37  & \cellcolor{Gold2!5!SeaGreen3!95} 44.50 & \cellcolor{Gold2!14!SeaGreen3!86} 63.81  & \cellcolor{Gold2!55!SeaGreen3!45} 0.0097 & \cellcolor{Gold2!32!SeaGreen3!68} 0.2115 \\
LLM+MT & Seed-X-PPO-7B & \cellcolor{Gold2!100!SeaGreen3!0} -9.68 & \cellcolor{Gold2!100!SeaGreen3!0} -6.32  & \cellcolor{Gold2!0!SeaGreen3!100} 53.18 & \cellcolor{Gold2!0!SeaGreen3!100} 71.99  & \cellcolor{Gold2!0!SeaGreen3!100} 45.64 & \cellcolor{Gold2!0!SeaGreen3!100} 67.19  & \cellcolor{Gold2!100!SeaGreen3!0} 0.0122 & \cellcolor{Gold2!19!SeaGreen3!81} 0.1665 \\
\midrule
& Average & \cellcolor{Gold2!19!SeaGreen3!81} -0.49 & \cellcolor{Gold2!21!SeaGreen3!79} -1.10 & \cellcolor{Gold2!49!SeaGreen3!51} 36.66 & \cellcolor{Gold2!45!SeaGreen3!55} 59.43  & \cellcolor{Gold2!37!SeaGreen3!63} 36.10 & \cellcolor{Gold2!34!SeaGreen3!66} 58.79  & \cellcolor{Gold2!41!SeaGreen3!59} 0.0089 & \cellcolor{Gold2!40!SeaGreen3!60} 0.2370 \\
\bottomrule
\end{tabular}
\caption{Aggregated results for each model: mean pairwise $\Delta$ scores; FLORES scores averaged across languages; \ours{} scores averaged across locales; and locale-specific word recalls averaged across locales.}
\label{tab:score by model}
\end{table*}

\section{Empirical Results and Analysis}

\subsection{Do \texorpdfstring{\ours{}}{} and FLORES yield consistent model rankings?}
We use the 32 models to individually translate FLORES and \ours{} test sets into English. We then compute BLEU and chrF scores for the translations using their respective references. Finally, we compute Spearman's and Kendall's rank correlation coefficients, $r_{s}$ and $\tau _{b}$, between model rankings for each language or locale, as determined by BLEU and chrF respectively.

\Cref{tab:ranking correlations} shows $r_s\!\geq\!0.93, \tau_b\!\geq\!0.80$ for all translation directions and $r_s\!\geq\!0.98, \tau_b\!\geq\!0.89$ on average. These indicate that \ours{} and FLORES counterparts induce highly consistent outcomes in ranking models. 

\begin{figure*}[t]
    \centering\small
    \begin{tikzpicture}
\begin{groupplot}[
    group style={
        group size=2 by 1,
        horizontal sep=0.07\textwidth,
    },
    width=0.52\textwidth,
    height=0.35\textwidth,
    xmode=log,
    log ticks with fixed point,
    xlabel={Total Parameters (B)},
    grid=none,
    axis line style={gray!80},
    axis lines*=left,
    ylabel shift=-5pt,
    legend style={
        at={(0.5, 0)},
        anchor=north,
        legend columns=8,
        draw=none,
        fill=none,
    },
    tick label style={font=\footnotesize},
    cycle list name=okabeito-accessible,
    every axis plot/.append style={thick}
]

\nextgroupplot[
    ylabel={$\overline{\Delta}_\text{BLEU}$},
    ymin=-2.5, ymax=2,
    ytick={-2,-1,0,1,2}
]
    \addplot+[
        visualization depends on={ \coordindex \as \myindex },
        nodes near coords={%
            \pgfmathparse{\myindex==3 ? "Llama 3.X" : ""}%
            \pgfmathresult%
        },
        every node near coord/.style={
            anchor=north, 
            font=\footnotesize, 
            scale=0.9,
            yshift=-1pt,
            xshift=-1pt,
        }
    ] coordinates { (1.0, 0.3) (3.0, 0.89) (8.0, 1.42) (70.0, -0.58) }; 
    
    \addplot+[
        visualization depends on={ \coordindex \as \myindex },
        nodes near coords={%
            \pgfmathparse{\myindex==6 ? "Qwen 3.5" : ""}%
            \pgfmathresult%
        },
        every node near coord/.style={
            anchor=north, 
            font=\footnotesize, 
            scale=0.9,
            yshift=-1pt,
            xshift=1pt,
        }
    ] coordinates { (0.8, -0.7) (2.0, 0.56) (4.0, 0.42) (9.0, 0.61) (27.0, 0.8) (35.0, 1.7) (122.0, 0.92) }; 

    \addplot+[
        visualization depends on={ \coordindex \as \myindex },
        nodes near coords={%
            \pgfmathparse{\myindex==0 ? "Gemma 3" : ""}%
            \pgfmathresult%
        },
        every node near coord/.style={
            anchor=north, 
            font=\footnotesize, 
            scale=0.9,
            yshift=-1pt,
            xshift=-9pt,
        }
    ] coordinates { (1.0, 0.27) (4.0, 0.96) }; 
    
    \addplot+[
        visualization depends on={ \coordindex \as \myindex },
        nodes near coords={%
            \pgfmathparse{\myindex==0 ? "Ministral" : ""}%
            \pgfmathresult%
        },
        every node near coord/.style={
            anchor=north, 
            font=\footnotesize, 
            scale=0.9,
            yshift=-3pt,
            xshift=0pt,
        }
    ] coordinates { (3.0, 0.52) (8.0, 1.4) }; 
    
    \addplot+[
        visualization depends on={ \coordindex \as \myindex },
        nodes near coords={%
            \pgfmathparse{\myindex==1 ? "EuroLLM" : ""}%
            \pgfmathresult%
        },
        every node near coord/.style={
            anchor=west, 
            font=\footnotesize, 
            scale=0.9,
            yshift=5pt,
            xshift=2pt,
        }
    ] coordinates { (1.7, 0.48) (9.0, 1.36) }; 
    
    \addplot+[
        visualization depends on={ \coordindex \as \myindex },
        nodes near coords={%
            \pgfmathparse{\myindex==1 ? "MADLAD" : ""}%
            \pgfmathresult%
        },
        every node near coord/.style={
            anchor=north, 
            font=\footnotesize, 
            scale=0.8,
            yshift=-2pt,
            xshift=1pt,
        }
    ] coordinates { (3.0, -0.9) (7.0, -0.53) }; 
    
    \addplot+[
        visualization depends on={ \coordindex \as \myindex },
        nodes near coords={%
            \pgfmathparse{\myindex==2 ? "Hy-MT2" : ""}%
            \pgfmathresult%
        },
        every node near coord/.style={
            anchor=south, 
            font=\footnotesize, 
            scale=0.9,
            xshift=2pt,
            yshift=1pt,
        }
    ] coordinates { (1.8, -2.06) (7.0, -2.16) (30.0, -2.28) }; 
    
    \addplot+[
        visualization depends on={ \coordindex \as \myindex },
        nodes near coords={%
            \pgfmathparse{\myindex==0 ? "NLLB" : ""}%
            \pgfmathresult%
        },
        every node near coord/.style={
            anchor=north, 
            font=\footnotesize, 
            scale=0.8,
            yshift=-2pt,
            xshift=0pt,
        }
    ] coordinates { (0.6, -1.21) (1.3, -0.79) }; 

\nextgroupplot[
    ylabel={$\overline{\Delta}_\text{chrF}$},
    ymin=-2.5, ymax=0.5,
    ytick={-2,-1.5,-1,-0.5,0,0.5},
    legend to name=grouplegend 
]
    \addplot+[
        visualization depends on={ \coordindex \as \myindex },
        nodes near coords={%
            \pgfmathparse{\myindex==3 ? "Llama 3.X" : ""}%
            \pgfmathresult%
        },
        every node near coord/.style={
            anchor=north, 
            font=\footnotesize, 
            scale=0.9,
            yshift=-1pt,
            xshift=-1.5pt,
        }
    ] coordinates { (1.0, -0.93) (3.0, -0.35) (8.0, 0.31) (70.0, -0.98) }; \addlegendentry{Llama 3.X}
    
    \addplot+[
        visualization depends on={ \coordindex \as \myindex },
        nodes near coords={%
            \pgfmathparse{\myindex==6 ? "Qwen 3.5" : ""}%
            \pgfmathresult%
        },
        every node near coord/.style={
            anchor=north, 
            font=\footnotesize, 
            scale=0.9,
            yshift=-1pt,
            xshift=-1pt,
        }
    ] coordinates { (0.8, -1.41) (2.0, -0.4) (4.0, -0.64) (9.0, -0.37) (27.0, -0.25) (35.0, 0.2) (122.0, -0.17) }; \addlegendentry{Qwen 3.5}
    
    \addplot+[
        visualization depends on={ \coordindex \as \myindex },
        nodes near coords={%
            \pgfmathparse{\myindex==0 ? "Gemma 3" : ""}%
            \pgfmathresult%
        },
        every node near coord/.style={
            anchor=south, 
            font=\footnotesize, 
            scale=0.9,
            yshift=1pt,
            xshift=-12pt,
        }
    ] coordinates { (1.0, -0.79) (4.0, -0.24) }; \addlegendentry{Gemma 3}
    
    \addplot+[
        visualization depends on={ \coordindex \as \myindex },
        nodes near coords={%
            \pgfmathparse{\myindex==1 ? "Ministral" : ""}%
            \pgfmathresult%
        },
        every node near coord/.style={
            anchor=north, 
            font=\footnotesize, 
            scale=0.9,
            yshift=-2.5pt,
            xshift=0pt,
        }
    ] coordinates { (3.0, -0.38) (8.0, -0.04) }; \addlegendentry{Ministral}
    
    \addplot+[
        visualization depends on={ \coordindex \as \myindex },
        nodes near coords={%
            \pgfmathparse{\myindex==0 ? "EuroLLM" : ""}%
            \pgfmathresult%
        },
        every node near coord/.style={
            anchor=south, 
            font=\footnotesize,
            scale=0.9,
            yshift=1.5pt,
            xshift=-10pt,
        }
    ] coordinates { (1.7, -0.43) (9.0, -0.03) }; \addlegendentry{EuroLLM}
    
    \addplot+[
        visualization depends on={ \coordindex \as \myindex },
        nodes near coords={%
            \pgfmathparse{\myindex==1 ? "MADLAD" : ""}%
            \pgfmathresult%
        },
        every node near coord/.style={
            anchor=north, 
            font=\footnotesize, 
            scale=0.8,
            yshift=-2pt,
            xshift=1pt,
        }
    ] coordinates { (3.0, -1.5) (7.0, -1.34) }; \addlegendentry{MADLAD}
    
    \addplot+[
        visualization depends on={ \coordindex \as \myindex },
        nodes near coords={%
            \pgfmathparse{\myindex==2 ? "Hy-MT2" : ""}%
            \pgfmathresult%
        },
        every node near coord/.style={
            anchor=south, 
            font=\footnotesize, 
            scale=0.9,
            xshift=2pt,
            yshift=1pt,
        }
    ] coordinates { (1.8, -2.09) (7.0, -2.06) (30.0, -2.21) }; \addlegendentry{Hy-MT2}
     
    \addplot+[
        visualization depends on={ \coordindex \as \myindex },
        nodes near coords={%
            \pgfmathparse{\myindex==0 ? "NLLB" : ""}%
            \pgfmathresult%
        },
        every node near coord/.style={
            anchor=north, 
            font=\footnotesize, 
            scale=0.8,
            yshift=-2pt,
            xshift=0pt,
        }
    ] coordinates { (0.6, -2.05) (1.3, -1.43) }; \addlegendentry{NLLB}

\end{groupplot}

\node at ($(group c1r1.south)!0.5!(group c2r1.south) - (0.2cm, 1.3cm)$) {\pgfplotslegendfromname{grouplegend}};

\end{tikzpicture}
    \vspace{-4ex}
    \caption{Plots of  $\Delta_\text{BLEU}$ (left) and  $\Delta_\text{chrf}$ (right) averaged across locales against model size (logarithmic scale).}
    \label{fig:delta against model size}
\end{figure*}
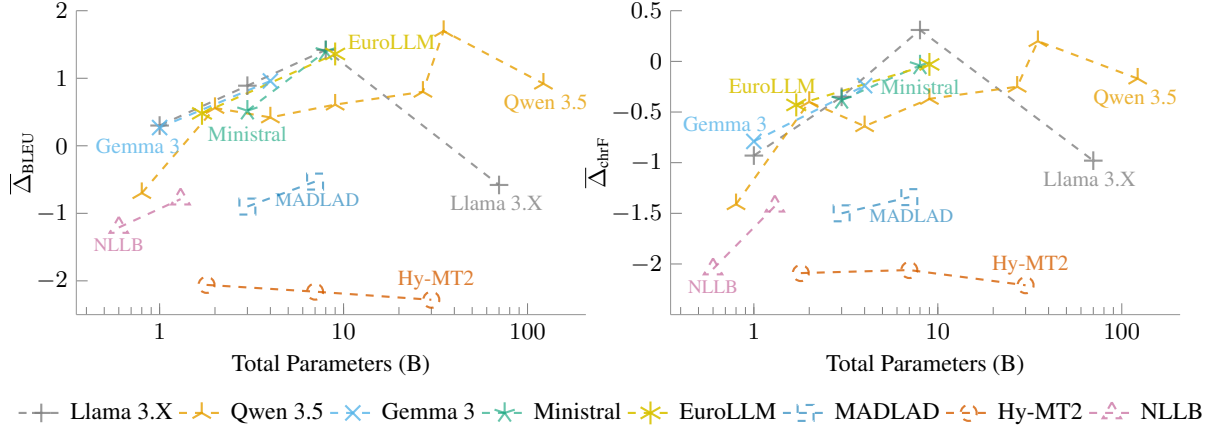

\subsection{Do models perform better or worse with localised source content?}

Our localised \ours{} demonstrates a more natural and aligned use case of source-to-English translation, since for each locale, all source instances are localised to relevant locations. To study the models' behaviour on the localised source, we report previously introduced metric scores in \Cref{tab:score by model}: 1) mean pairwise $\Delta_\text{BLEU}$ and $\Delta_\text{chrF}$; 2) BLEU and chrF on FLORES; 3) BLEU and chrF on \ours{}; and 4) false positive and false negative rates for unique word recall. It is worth noting that a mean pairwise $\Delta$ score does not equal the difference between the averaged scores for FLORES and \ours{}, because one language can be paired with multiple locales.

As \Cref{tab:score by model} shows, for most models, $\Delta_\text{BLEU}$ ranges between $-2$ and $+2$, and $\Delta_\text{chrF}$ ranges between $-2$ and $+0.7$, with averages of $-0.49$ and $-1.1$, respectively. Specifically, \texttt{SmolLM-1.7B-Instruct} acts as a ``random'' baseline---as the model is not good at translation, reflected by its BLEU and chrF---with $\Delta$ scores close to $0$. However, two models at the bottom see significantly lower $\Delta$ scores when the input is localised, with a $>\!8$ BLEU drop and a $>\!5$ chrF drop.

Most models' false negatives fall within 10--30\%, with a few higher numbers associated with weaker translation models showing low BLEU and chrF scores, which is largely expected. If we treat these weaker models with $\text{FN}\!>$44\% as random baselines, we can conclude that most models can adapt to the localised source to a certain extent.

\paragraph{Model type} Considering how models have been trained, MT-optimised ones rank lower by $\Delta$ scores, including both pre-trained translation models (\texttt{NLLB,MADLAD}) and LLMs further optimised for translation (\texttt{Hy-MT2,Seed-X-PPO-7B}). Even though some of these have a small magnitude, they are consistently clustered in the negative region.

\paragraph{Model size} We then plot  $\Delta_\text{BLEU}$ and average $\Delta_\text{chrf}$ against model size in \Cref{fig:delta against model size} using different marks and colours to represent model families. In general, we see $\Delta_\text{BLEU}$ and $\Delta_\text{chrf}$ increase as the model gets larger for 6 out of 8 families. Two families do not follow this pattern: \texttt{Hy-MT2} slightly drops as it gets larger, and \texttt{Llama\,3.X} peaks at 8B followed by a sharp decline at 70B---interpreted with reservations, because \texttt{Llama} spans version 3.1--3.3. Since $\Delta$ measures the performance difference between FLORES and our \ours{}, an in-family scaling of $\Delta$ means that the BLEU scores on the two test sets diverge as models get larger. This could imply that FLORES may underestimate the true multilingual translation capabilities of stronger models.

\begin{figure*}[t!]
\centering\small
\begin{tikzpicture}
\begin{groupplot}[
    group style={
        group size=2 by 1,
        horizontal sep=1.2cm
    },
    width=0.5\textwidth,
    height=0.3\textwidth,
    ybar=0.05pt,
    /pgf/bar width=8pt,
    unbounded coords=jump,
    symbolic x coords={US, IN, DE, UK, CN, SG},
    xtick={US, IN, DE, UK, CN, SG},
    x tick label style={
        rotate=30,
        anchor=north east,
        yshift=-0.5pt,
        xshift=2.2pt,
        font=\footnotesize,
        scale=0.8,
        inner sep=0.5pt
    },
    y tick label style={font=\footnotesize},
    title style={font=\footnotesize, yshift=-1ex},
    enlarge x limits=0.15,
    grid=none,
    axis line style={gray!80},
    tick pos=left,
    extra y ticks={0},
    extra y tick labels={},
    extra y tick style={
        grid=major,
        major grid style={dashed, draw=gray!80}
    }
]

\nextgroupplot[
    title={$\overline{\Delta}_\text{BLEU}$ by Location},
    legend style={
        at={(0.15,0.35)},
        anchor=north,
        font=\footnotesize,
        draw=none,
        fill=white,
        fill opacity=1,
        text opacity=1,
        inner sep=-1pt,
        row sep=-1pt,
        text width=1.5em,
        align=left,
        legend cell align=left,
    },
    legend image code/.code={
        \draw[
            fill=#1,
            draw=white
        ] (0cm,-0.12cm) rectangle (0.35cm,0.22cm);
    },
]

\addlegendimage{fill=oiBluishGreen}
\addlegendentry{Hindi}

\addlegendimage{fill=oiVermilion}
\addlegendentry{Chinese}

\addplot[
    fill=oiBluishGreen,
    draw=none,
]
coordinates {
    (US,1.38)
    (IN,1.00)
    (DE,-0.28)
    (UK,nan)
    (CN,nan)
    (SG,nan)
};

\addplot[
    fill=oiVermilion,
    draw=none,
]
coordinates {
    (US,0.92)
    (IN,nan)
    (DE,nan)
    (UK,-1.03)
    (CN,-1.51)
    (SG,-4.29)
};

\nextgroupplot[
    title={$\overline{\Delta}_\text{chrF}$ by Location},
]

\addplot[
    fill=oiBluishGreen,
    draw=none,
]
coordinates {
    (US,1.36)
    (IN,-0.76)
    (DE,-1.31)
    (UK,nan)
    (CN,nan)
    (SG,nan)
};

\addplot[
    fill=oiVermilion,
    draw=none,
]
coordinates {
    (US,1.11)
    (IN,nan)
    (DE,nan)
    (UK,-0.52)
    (CN,-1.06)
    (SG,-2.22)
};

\end{groupplot}
\end{tikzpicture}
\caption{Plots of $\Delta_\text{BLEU}$ and $\Delta_\text{chrF}$ averaged across models for translating localised Hindi and Chinese to English.}
\label{fig:histogram deltas language different locations}
\end{figure*}
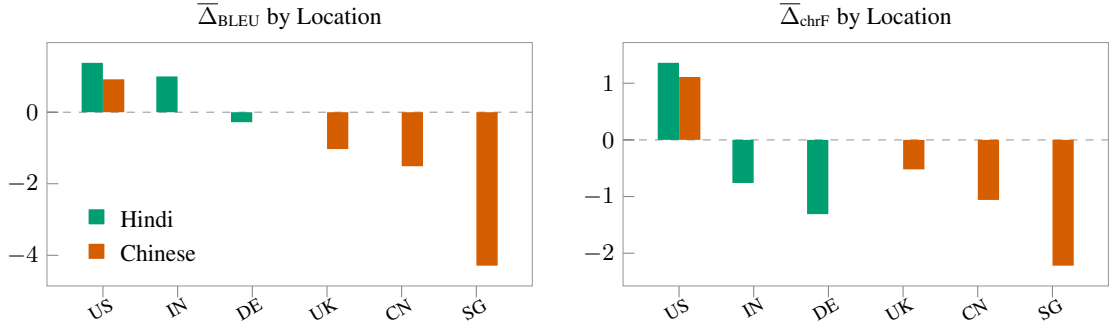

\subsection{Are models FLORES contaminated?}

Looking at BLEU and chrF scores, \texttt{aya-expanse-8b} and \texttt{Seed-X-PPO-7B} are strong on both FLORES and \ours{} tests, beating much larger general-purpose LLMs like \texttt{Llama-3.3-70B-Instruct} and \texttt{Qwen3.5-122B-A10B-FP8}. However, they exhibit a much larger performance decline ($\Delta_\text{BLEU}\!<\!-8$) than other models ($-2\!<\!\Delta_\text{BLEU}\!<\!+2$) when moving from FLORES to \ours{}. A further score breakdown in \Cref{sec:delta breakdown} \Cref{fig:breakdown of deltas} shows that, for the two models, a substantial BLEU or chrF drop occurs in most source-to-English directions, implying a model artifact.

We further test for text memorization---whether verbatim FLORES English words are undesirably recalled when translating counterpart \ours{} instances. We probe this via false positives as defined in \Cref{subsec:metrics} and reported in \Cref{tab:score by model}. Most models fall below 1\%. Looking at the models at the bottom of the table with negative $\Delta$ scores, they do not record significantly higher false positives compared to models at the top. Hence, there is no direct evidence that these models have memorised the exact FLORES text. Nonetheless, disproportionate drops in BLEU and chrF could hint at a lack of robustness to localisation or some degree of overfitting to FLORES, since localisation still retains the original sentence structure. This is corroborated by an error analysis in \Cref{sec:manual-analysis-findings} later.

\begin{table}[t!]
\centering\small
\setlength{\tabcolsep}{1.2ex}
\begin{tabular}{lllcc}
\toprule
\textbf{Lang} & \textbf{Script} & \textbf{Location} & \textbf{$\overline{\Delta}_\text{BLEU}\!\downarrow$} & \textbf{$\overline{\Delta}_\text{chrF}$} \\
\midrule
plt & Latn & Madagascar & \cellcolor{Gold2!0!SeaGreen3!100} \phantom{-}2.53 & \cellcolor{Gold2!12!SeaGreen3!88} \phantom{-}0.86 \\
gom & Deva & India & \cellcolor{Gold2!7!SeaGreen3!93} \phantom{-}2.05 & \cellcolor{Gold2!33!SeaGreen3!67} -0.06 \\
yor & Latn & Nigeria & \cellcolor{Gold2!15!SeaGreen3!85} \phantom{-}1.48 & \cellcolor{Gold2!30!SeaGreen3!70} \phantom{-}0.09 \\
hin & Deva & US & \cellcolor{Gold2!17!SeaGreen3!83} \phantom{-}1.38 & \cellcolor{Gold2!0!SeaGreen3!100} \phantom{-}1.36 \\
tel & Telu & India & \cellcolor{Gold2!20!SeaGreen3!80} \phantom{-}1.14 & \cellcolor{Gold2!39!SeaGreen3!61} -0.27 \\
hin & Deva & India & \cellcolor{Gold2!22!SeaGreen3!78} \phantom{-}1.00 & \cellcolor{Gold2!50!SeaGreen3!50} -0.76 \\
ben & Beng & India & \cellcolor{Gold2!23!SeaGreen3!77} \phantom{-}0.96 & \cellcolor{Gold2!41!SeaGreen3!59} -0.38 \\
cmn & Hans & US & \cellcolor{Gold2!24!SeaGreen3!76} \phantom{-}0.92 & \cellcolor{Gold2!6!SeaGreen3!94} \phantom{-}1.11 \\
pol & Latn & Poland & \cellcolor{Gold2!31!SeaGreen3!69} \phantom{-}0.38 & \cellcolor{Gold2!77!SeaGreen3!23} -1.89 \\
spa & Latn & Spain & \cellcolor{Gold2!36!SeaGreen3!64} \phantom{-}0.05 & \cellcolor{Gold2!50!SeaGreen3!50} -0.76 \\
deu & Latn & Germany & \cellcolor{Gold2!37!SeaGreen3!63} \phantom{-}0.02 & \cellcolor{Gold2!62!SeaGreen3!38} -1.25 \\
bul & Cyrl & Bulgaria & \cellcolor{Gold2!40!SeaGreen3!60} -0.21 & \cellcolor{Gold2!47!SeaGreen3!53} -0.65 \\
hin & Deva & Germany & \cellcolor{Gold2!41!SeaGreen3!59} -0.28 & \cellcolor{Gold2!63!SeaGreen3!37} -1.31 \\
slk & Latn & Slovakia & \cellcolor{Gold2!41!SeaGreen3!59} -0.30 & \cellcolor{Gold2!69!SeaGreen3!31} -1.55 \\
fil & Latn & Philippines & \cellcolor{Gold2!47!SeaGreen3!53} -0.68 & \cellcolor{Gold2!45!SeaGreen3!55} -0.57 \\
tur & Latn & Turkey & \cellcolor{Gold2!51!SeaGreen3!49} -0.96 & \cellcolor{Gold2!77!SeaGreen3!23} -1.91 \\
cmn & Hans & UK & \cellcolor{Gold2!52!SeaGreen3!48} -1.03 & \cellcolor{Gold2!44!SeaGreen3!56} -0.52 \\
cat & Latn & Spain & \cellcolor{Gold2!55!SeaGreen3!45} -1.24 & \cellcolor{Gold2!62!SeaGreen3!38} -1.26 \\
ces & Latn & Czechia & \cellcolor{Gold2!56!SeaGreen3!44} -1.30 & \cellcolor{Gold2!83!SeaGreen3!17} -2.16 \\
cmn & Hans & China & \cellcolor{Gold2!59!SeaGreen3!41} -1.51 & \cellcolor{Gold2!57!SeaGreen3!43} -1.06 \\
rus & Cyrl & Russia & \cellcolor{Gold2!62!SeaGreen3!38} -1.70 & \cellcolor{Gold2!75!SeaGreen3!25} -1.80 \\
deu & Latn & Switzerland & \cellcolor{Gold2!66!SeaGreen3!34} -1.97 & \cellcolor{Gold2!87!SeaGreen3!13} -2.31 \\
acm & Arab & Iraq & \cellcolor{Gold2!67!SeaGreen3!33} -2.06 & \cellcolor{Gold2!87!SeaGreen3!13} -2.32 \\
khk & Cyrl & Mongolia & \cellcolor{Gold2!68!SeaGreen3!32} -2.13 & \cellcolor{Gold2!100!SeaGreen3!0} -2.88 \\
jpn & Jpan & Japan & \cellcolor{Gold2!76!SeaGreen3!24} -2.65 & \cellcolor{Gold2!98!SeaGreen3!2} -2.81 \\
apc & Arab & Syria & \cellcolor{Gold2!78!SeaGreen3!22} -2.77 & \cellcolor{Gold2!91!SeaGreen3!9} -2.49 \\
cmn & Hans & Singapore & \cellcolor{Gold2!100!SeaGreen3!0} -4.29 & \cellcolor{Gold2!84!SeaGreen3!16} -2.22 \\
\midrule
\multicolumn{3}{l}{Average} & \cellcolor{Gold2!41!SeaGreen3!59} -0.27 & \cellcolor{Gold2!57!SeaGreen3!43} -1.05 \\
\bottomrule
\end{tabular}
\caption{Average $\Delta_\text{BLEU}$ and $\Delta_\text{chrF}$ for each locale.}
\label{tab:tab:average deltas across models}
\end{table}

\subsection{Which locales are harder?}  

\Cref{tab:tab:average deltas across models} lists $\Delta_\text{BLEU}$ and $\Delta_\text{chrF}$ scores again, but this time averaged across models for each locale. Most $\Delta_\text{BLEU}$ range from $-2$ to $+2$, and most chrF range from $-2$ to $+1$. According to BLEU differences, models perform better on localised \texttt{plt\_Latn\_Madagascar} and \texttt{gom\_Deva\_India}; according to chrF differences, models perform better on \texttt{hin\_Deva\_US}, \texttt{cmn\_Hans\_US}, and \texttt{plt\_Latn\_Madagascar}. At the bottom of the table, \texttt{cmn\_Hans\_Singapore} receives significantly low $\Delta$, and another five localised test sets also have $\Delta_\text{BLEU}$ and  $\Delta_\text{chrF}$ around $-2$. 

These results demonstrate that the difficulty of a test set could change after being localised---some appear easier while others become harder. Nonetheless, it is noteworthy that localisations were produced by an LLM and edited by individual annotators, who may bring in their personal biases, so the difficulty of individual locales could be attributed to a combination of model artifact, e.g. training data, as well as data artifact relative to the original FLORES instances.

\begin{table*}[t!]
\centering\small
\setlength{\tabcolsep}{0.1em}

\begin{tabular}{l @{\hspace{0.4em}} ccccccc @{\hspace{0.6em}} ccccccc @{\hspace{0.6em}} ccccccc}
\toprule
\multicolumn{1}{c}{\multirow[b]{3}{*}{Model}}& \multicolumn{7}{c}{Chinese\_CN$\to$English}
& \multicolumn{7}{c}{Chinese\_UK$\to$English}
& \multicolumn{7}{c}{Chinese\_US$\to$English} 
\\[-2pt]
\cmidrule(lr){2-8}
\cmidrule(lr){9-15}
\cmidrule(lr){16-22}
& \multicolumn{4}{c}{Locale}
& \multirow[b]{2}{*}{\rotatebox[origin=b]{60}{\phantom{\,\,}General}}
& \multirow[b]{2}{*}{\rotatebox[origin=b]{60}{Total}}
& \multirow[b]{2}{*}{\rotatebox[origin=b]{60}{\phantom{\,\,}BLEU}}
& \multicolumn{4}{c}{Locale}
& \multirow[b]{2}{*}{\rotatebox[origin=b]{60}{\phantom{\,\,}General}}
& \multirow[b]{2}{*}{\rotatebox[origin=b]{60}{Total}}
& \multirow[b]{2}{*}{\rotatebox[origin=b]{60}{\phantom{\,\,}BLEU}}
& \multicolumn{4}{c}{Locale}
& \multirow[b]{2}{*}{\rotatebox[origin=b]{60}{\phantom{\,\,}General}}
& \multirow[b]{2}{*}{\rotatebox[origin=b]{60}{Total}}
& \multirow[b]{2}{*}{\rotatebox[origin=b]{60}{\phantom{\,\,}BLEU}}
\\
\cmidrule{2-5}
\cmidrule{9-12}
\cmidrule{16-19}
& {NE} & {PY} & {IC} & {NU} &  &  & 
& {NE} & {PY} & {IC} & {NU} &  &  & 
& {NE} & {PY} & {IC} & {NU} &  &  & 
\\
\midrule
Seed-X-PPO-7B & \cellcolor{Gold2!14!SeaGreen3!86} 8 & \cellcolor{Gold2!70!SeaGreen3!30} 7 & \cellcolor{Gold2!33!SeaGreen3!67} 1 & \cellcolor{Gold2!20!SeaGreen3!80} 1 & \cellcolor{Gold2!11!SeaGreen3!89} 4 & \cellcolor{Gold2!26!SeaGreen3!74} 21 & \cellcolor{Gold2!33!SeaGreen3!67} 41.87 & \cellcolor{Gold2!25!SeaGreen3!75} 12 & \cellcolor{Gold2!0!SeaGreen3!100} 0 & \cellcolor{Gold2!0!SeaGreen3!100} 0 & \cellcolor{Gold2!20!SeaGreen3!80} 1 & \cellcolor{Gold2!33!SeaGreen3!67} 8 & \cellcolor{Gold2!26!SeaGreen3!74} 21 & \cellcolor{Gold2!14!SeaGreen3!86} 44.94 & \cellcolor{Gold2!28!SeaGreen3!72} 13 & \cellcolor{Gold2!0!SeaGreen3!100} 0 & \cellcolor{Gold2!33!SeaGreen3!67} 1 & \cellcolor{Gold2!20!SeaGreen3!80} 1 & \cellcolor{Gold2!11!SeaGreen3!89} 4 & \cellcolor{Gold2!23!SeaGreen3!77} 19 & \cellcolor{Gold2!0!SeaGreen3!100} 47.31 \\
Llama-3.1-8B\tiny{-Instruct} & \cellcolor{Gold2!33!SeaGreen3!67} 15 & \cellcolor{Gold2!70!SeaGreen3!30} 7 & \cellcolor{Gold2!33!SeaGreen3!67} 1 & \cellcolor{Gold2!40!SeaGreen3!60} 2 & \cellcolor{Gold2!56!SeaGreen3!44} 12 & \cellcolor{Gold2!57!SeaGreen3!43} 37 & \cellcolor{Gold2!85!SeaGreen3!15} 33.45 & \cellcolor{Gold2!56!SeaGreen3!44} 23 & \cellcolor{Gold2!0!SeaGreen3!100} 0 & \cellcolor{Gold2!33!SeaGreen3!67} 1 & \cellcolor{Gold2!100!SeaGreen3!0} 5 & \cellcolor{Gold2!56!SeaGreen3!44} 12 & \cellcolor{Gold2!64!SeaGreen3!36} 41 & \cellcolor{Gold2!81!SeaGreen3!19} 34.00 & \cellcolor{Gold2!36!SeaGreen3!64} 16 & \cellcolor{Gold2!0!SeaGreen3!100} 0 & \cellcolor{Gold2!0!SeaGreen3!100} 0 & \cellcolor{Gold2!20!SeaGreen3!80} 1 & \cellcolor{Gold2!28!SeaGreen3!72} 7 & \cellcolor{Gold2!32!SeaGreen3!68} 24 & \cellcolor{Gold2!73!SeaGreen3!27} 35.36 \\
Ministral-3-8B\tiny{-Instruct-2512} & \cellcolor{Gold2!67!SeaGreen3!33} 27 & \cellcolor{Gold2!70!SeaGreen3!30} 7 & \cellcolor{Gold2!67!SeaGreen3!33} 2 & \cellcolor{Gold2!60!SeaGreen3!40} 3 & \cellcolor{Gold2!50!SeaGreen3!50} 11 & \cellcolor{Gold2!81!SeaGreen3!19} 50 & \cellcolor{Gold2!91!SeaGreen3!9} 32.32 & \cellcolor{Gold2!58!SeaGreen3!42} 24 & \cellcolor{Gold2!0!SeaGreen3!100} 0 & \cellcolor{Gold2!0!SeaGreen3!100} 0 & \cellcolor{Gold2!60!SeaGreen3!40} 3 & \cellcolor{Gold2!67!SeaGreen3!33} 14 & \cellcolor{Gold2!64!SeaGreen3!36} 41 & \cellcolor{Gold2!100!SeaGreen3!0} 30.92 & \cellcolor{Gold2!42!SeaGreen3!58} 18 & \cellcolor{Gold2!0!SeaGreen3!100} 0 & \cellcolor{Gold2!0!SeaGreen3!100} 0 & \cellcolor{Gold2!20!SeaGreen3!80} 1 & \cellcolor{Gold2!28!SeaGreen3!72} 7 & \cellcolor{Gold2!36!SeaGreen3!64} 26 & \cellcolor{Gold2!83!SeaGreen3!17} 33.63 \\
EuroLLM-9B\tiny{-Instruct-2512} & \cellcolor{Gold2!36!SeaGreen3!64} 16 & \cellcolor{Gold2!100!SeaGreen3!0} 10 & \cellcolor{Gold2!67!SeaGreen3!33} 2 & \cellcolor{Gold2!40!SeaGreen3!60} 2 & \cellcolor{Gold2!17!SeaGreen3!83} 5 & \cellcolor{Gold2!53!SeaGreen3!47} 35 & \cellcolor{Gold2!58!SeaGreen3!42} 37.81 & \cellcolor{Gold2!36!SeaGreen3!64} 16 & \cellcolor{Gold2!0!SeaGreen3!100} 0 & \cellcolor{Gold2!0!SeaGreen3!100} 0 & \cellcolor{Gold2!40!SeaGreen3!60} 2 & \cellcolor{Gold2!56!SeaGreen3!44} 12 & \cellcolor{Gold2!43!SeaGreen3!57} 30 & \cellcolor{Gold2!62!SeaGreen3!38} 37.22 & \cellcolor{Gold2!28!SeaGreen3!72} 13 & \cellcolor{Gold2!0!SeaGreen3!100} 0 & \cellcolor{Gold2!0!SeaGreen3!100} 0 & \cellcolor{Gold2!0!SeaGreen3!100} 0 & \cellcolor{Gold2!17!SeaGreen3!83} 5 & \cellcolor{Gold2!21!SeaGreen3!79} 18 & \cellcolor{Gold2!52!SeaGreen3!48} 38.82 \\
Qwen3.5-2B & \cellcolor{Gold2!83!SeaGreen3!17} 33 & \cellcolor{Gold2!70!SeaGreen3!30} 7 & \cellcolor{Gold2!100!SeaGreen3!0} 3 & \cellcolor{Gold2!80!SeaGreen3!20} 4 & \cellcolor{Gold2!61!SeaGreen3!39} 13 & \cellcolor{Gold2!100!SeaGreen3!0} 60 & \cellcolor{Gold2!89!SeaGreen3!11} 32.79 & \cellcolor{Gold2!100!SeaGreen3!0} 39 & \cellcolor{Gold2!0!SeaGreen3!100} 0 & \cellcolor{Gold2!0!SeaGreen3!100} 0 & \cellcolor{Gold2!100!SeaGreen3!0} 5 & \cellcolor{Gold2!72!SeaGreen3!28} 15 & \cellcolor{Gold2!98!SeaGreen3!2} 59 & \cellcolor{Gold2!94!SeaGreen3!6} 31.89 & \cellcolor{Gold2!56!SeaGreen3!44} 23 & \cellcolor{Gold2!0!SeaGreen3!100} 0 & \cellcolor{Gold2!0!SeaGreen3!100} 0 & \cellcolor{Gold2!40!SeaGreen3!60} 2 & \cellcolor{Gold2!100!SeaGreen3!0} 20 & \cellcolor{Gold2!72!SeaGreen3!28} 45 & \cellcolor{Gold2!81!SeaGreen3!19} 34.02 \\
Qwen3.5-4B & \cellcolor{Gold2!25!SeaGreen3!75} 12 & \cellcolor{Gold2!60!SeaGreen3!40} 6 & \cellcolor{Gold2!0!SeaGreen3!100} 0 & \cellcolor{Gold2!20!SeaGreen3!80} 1 & \cellcolor{Gold2!28!SeaGreen3!72} 7 & \cellcolor{Gold2!36!SeaGreen3!64} 26 & \cellcolor{Gold2!72!SeaGreen3!28} 35.59 & \cellcolor{Gold2!58!SeaGreen3!42} 24 & \cellcolor{Gold2!0!SeaGreen3!100} 0 & \cellcolor{Gold2!0!SeaGreen3!100} 0 & \cellcolor{Gold2!40!SeaGreen3!60} 2 & \cellcolor{Gold2!44!SeaGreen3!56} 10 & \cellcolor{Gold2!55!SeaGreen3!45} 36 & \cellcolor{Gold2!73!SeaGreen3!27} 35.41 & \cellcolor{Gold2!39!SeaGreen3!61} 17 & \cellcolor{Gold2!0!SeaGreen3!100} 0 & \cellcolor{Gold2!0!SeaGreen3!100} 0 & \cellcolor{Gold2!0!SeaGreen3!100} 0 & \cellcolor{Gold2!22!SeaGreen3!78} 6 & \cellcolor{Gold2!30!SeaGreen3!70} 23 & \cellcolor{Gold2!71!SeaGreen3!29} 35.74 \\
Qwen3.5-9B & \cellcolor{Gold2!19!SeaGreen3!81} 10 & \cellcolor{Gold2!30!SeaGreen3!70} 3 & \cellcolor{Gold2!33!SeaGreen3!67} 1 & \cellcolor{Gold2!0!SeaGreen3!100} 0 & \cellcolor{Gold2!28!SeaGreen3!72} 7 & \cellcolor{Gold2!26!SeaGreen3!74} 21 & \cellcolor{Gold2!61!SeaGreen3!39} 37.32 & \cellcolor{Gold2!31!SeaGreen3!69} 14 & \cellcolor{Gold2!0!SeaGreen3!100} 0 & \cellcolor{Gold2!0!SeaGreen3!100} 0 & \cellcolor{Gold2!20!SeaGreen3!80} 1 & \cellcolor{Gold2!17!SeaGreen3!83} 5 & \cellcolor{Gold2!25!SeaGreen3!75} 20 & \cellcolor{Gold2!62!SeaGreen3!38} 37.23 & \cellcolor{Gold2!17!SeaGreen3!83} 9 & \cellcolor{Gold2!0!SeaGreen3!100} 0 & \cellcolor{Gold2!0!SeaGreen3!100} 0 & \cellcolor{Gold2!0!SeaGreen3!100} 0 & \cellcolor{Gold2!11!SeaGreen3!89} 4 & \cellcolor{Gold2!11!SeaGreen3!89} 13 & \cellcolor{Gold2!56!SeaGreen3!44} 38.10 \\
Qwen3.5-27B & \cellcolor{Gold2!3!SeaGreen3!97} 4 & \cellcolor{Gold2!20!SeaGreen3!80} 2 & \cellcolor{Gold2!0!SeaGreen3!100} 0 & \cellcolor{Gold2!0!SeaGreen3!100} 0 & \cellcolor{Gold2!0!SeaGreen3!100} 2 & \cellcolor{Gold2!2!SeaGreen3!98} 8 & \cellcolor{Gold2!55!SeaGreen3!45} 38.30 & \cellcolor{Gold2!17!SeaGreen3!83} 9 & \cellcolor{Gold2!0!SeaGreen3!100} 0 & \cellcolor{Gold2!0!SeaGreen3!100} 0 & \cellcolor{Gold2!40!SeaGreen3!60} 2 & \cellcolor{Gold2!0!SeaGreen3!100} 2 & \cellcolor{Gold2!11!SeaGreen3!89} 13 & \cellcolor{Gold2!57!SeaGreen3!43} 38.01 & \cellcolor{Gold2!6!SeaGreen3!94} 5 & \cellcolor{Gold2!0!SeaGreen3!100} 0 & \cellcolor{Gold2!0!SeaGreen3!100} 0 & \cellcolor{Gold2!0!SeaGreen3!100} 0 & \cellcolor{Gold2!11!SeaGreen3!89} 4 & \cellcolor{Gold2!4!SeaGreen3!96} 9 & \cellcolor{Gold2!48!SeaGreen3!52} 39.44 \\
Qwen3.5-35B-A3B & \cellcolor{Gold2!6!SeaGreen3!94} 5 & \cellcolor{Gold2!30!SeaGreen3!70} 3 & \cellcolor{Gold2!0!SeaGreen3!100} 0 & \cellcolor{Gold2!0!SeaGreen3!100} 0 & \cellcolor{Gold2!0!SeaGreen3!100} 2 & \cellcolor{Gold2!6!SeaGreen3!94} 10 & \cellcolor{Gold2!57!SeaGreen3!43} 38.00 & \cellcolor{Gold2!8!SeaGreen3!92} 6 & \cellcolor{Gold2!0!SeaGreen3!100} 0 & \cellcolor{Gold2!0!SeaGreen3!100} 0 & \cellcolor{Gold2!20!SeaGreen3!80} 1 & \cellcolor{Gold2!11!SeaGreen3!89} 4 & \cellcolor{Gold2!8!SeaGreen3!92} 11 & \cellcolor{Gold2!61!SeaGreen3!39} 37.29 & \cellcolor{Gold2!14!SeaGreen3!86} 8 & \cellcolor{Gold2!0!SeaGreen3!100} 0 & \cellcolor{Gold2!0!SeaGreen3!100} 0 & \cellcolor{Gold2!0!SeaGreen3!100} 0 & \cellcolor{Gold2!6!SeaGreen3!94} 3 & \cellcolor{Gold2!8!SeaGreen3!92} 11 & \cellcolor{Gold2!55!SeaGreen3!45} 38.28 \\
Qwen3.5-122B-A10B\tiny{-FP8} & \cellcolor{Gold2!0!SeaGreen3!100} 3 & \cellcolor{Gold2!20!SeaGreen3!80} 2 & \cellcolor{Gold2!0!SeaGreen3!100} 0 & \cellcolor{Gold2!20!SeaGreen3!80} 1 & \cellcolor{Gold2!6!SeaGreen3!94} 3 & \cellcolor{Gold2!4!SeaGreen3!96} 9 & \cellcolor{Gold2!54!SeaGreen3!46} 38.48 & \cellcolor{Gold2!3!SeaGreen3!97} 4 & \cellcolor{Gold2!0!SeaGreen3!100} 0 & \cellcolor{Gold2!0!SeaGreen3!100} 0 & \cellcolor{Gold2!0!SeaGreen3!100} 0 & \cellcolor{Gold2!6!SeaGreen3!94} 3 & \cellcolor{Gold2!0!SeaGreen3!100} 7 & \cellcolor{Gold2!59!SeaGreen3!41} 37.59 & \cellcolor{Gold2!17!SeaGreen3!83} 9 & \cellcolor{Gold2!0!SeaGreen3!100} 0 & \cellcolor{Gold2!0!SeaGreen3!100} 0 & \cellcolor{Gold2!0!SeaGreen3!100} 0 & \cellcolor{Gold2!0!SeaGreen3!100} 2 & \cellcolor{Gold2!8!SeaGreen3!92} 11 & \cellcolor{Gold2!53!SeaGreen3!47} 38.63 \\
\midrule
Total & 133 & 54 & 10 & 14 & 66 & 277 & & 171 & 0 & 1 & 22 & 85 & 279 & & 131 & 0 & 1 & 5 & 62 & 199 & \\
\bottomrule
\end{tabular}
\caption{Error counts and BLEU by model and country. We list locale-specific, general, and total errors separately. Locale-specific errors include named entities (NE), Pinyin (PY), idiom and culture (IC), and number and unit (NU).}
\label{tab:full-error-table}
\end{table*}

\subsection{Same language, different localisations}

In our test set, Hindi and Chinese have been localised to locations where they are not the official or de facto language, allowing us to conduct a small-scale ablation study on how models respond to localisation while controlling the language. In detail, Hindi has been localised to the US and Germany in addition to India; Chinese has been localised to the US and UK in addition to China and Singapore.

In \Cref{fig:histogram deltas language different locations}, we visualise the $\Delta_\text{BLEU}$ and $\Delta_\text{chrF}$ for both languages, averaged across models but separated by location. We see negative $\Delta$ for Germany, China, and Singapore, mixed results for India, and positive $\Delta$s for the US. Moreover, since the $\Delta$ scores are relative to FLORES, we can compare the performance between two localisations directly by comparing $\Delta$s. For example, when translating from Chinese to English, by definition: 
\begin{equation}
\begin{split}
\text{BLEU}_{\text{\texttt{cmn}\_US}} - \text{BLEU}_{\text{\texttt{cmn}\_SG}}=\\{\Delta_\text{BLEU}}_{\text{\texttt{cmn}\_US}} - {\Delta_\text{BLEU}}_{\text{\texttt{cmn}\_SG}}\,\text{.}\nonumber
\end{split}
\end{equation}
A positive ${\Delta_\text{BLEU}}_{\text{\texttt{cmn}\_US}}$ and a negative ${\Delta_\text{BLEU}}_{\text{\texttt{cmn}\_SG}}$ in \Cref{fig:histogram deltas language different locations} suggests that the difference between $\text{BLEU}_{\text{\texttt{cmn}\_US}}$ and $\text{BLEU}_{\text{\texttt{cmn}\_SG}}$ is significant. The same applies to chrF scores too.

Overall, the results indicate that, relative to the original FLORES, models translate US content better in both Hindi and Chinese, but translate content about China and Singapore worse in Chinese. The positive $\Delta$ for the US implies that, compared to FLORES, \ours{}'s US localisation is easier for models, even though the source languages, Hindi and Chinese, are not de facto in the US. On the other hand, the negative $\Delta$s are not ideal, because they indicate that models are less capable of translating source-localised content into English, which is the main use case of source-to-English translation. From a model building perspective, this contrastive evaluation reflects a US-centric bias in training data, and \ours{} has provided a systematic way to probe such locale-centric bias.

\section{Manual Error Analysis}

\subsection{Setup}
In addition to analysis based on automatic evaluation results, we conduct a manual inspection of translations of regionally localised content. In particular, a Chinese native speaker who is fluent in English inspected the Chinese-to-English outputs from selected models together with sources and references. 

Nine error types are pre-defined and grouped into two categories:
\begin{itemize}[itemsep=0.3ex,topsep=0.3ex]
    \item \textbf{Four locale-specific errors:} named entities, idioms and culturally specific expressions, number or unit, as well as Pinyin, a deterministic phonetic conversion system for transcribing Mandarin Chinese names into the Latin alphabet.
    \item \textbf{Five general errors:} semantic mistranslation, omitted information, hallucinated or added information, original language retained, and unwanted reasoning or formatting tokens.
\end{itemize}
We classify errors based on their manifestation rather than underlying cause, with errors directly concerning localised content categorised as locale-specific. An observed error may nevertheless arise from either locale-specific or general difficulties. Repeated errors of the same type within an instance are counted only once, while different types are counted separately.

\subsection{Observations corroborate earlier findings}\label{sec:manual-analysis-findings}

\Cref{tab:full-error-table} presents the results of our human error analysis, and we also include BLEU for reference. Across the three regional variants, Chinese content grounded in the US yields the fewest errors, with this difference becoming more pronounced among smaller models with higher error rates. 

Across models, we observe a clear scaling trend: locale-specific (and general) errors decrease as the model's total parameter count increases, while the number of active parameters plays a minor role. This pattern is evident in the Qwen3.5 mixture-of-experts models: both \texttt{35B-A3B} and \texttt{122B-A10B} exhibit substantially fewer errors than the \texttt{2B,4B,9B} dense models, despite activating a smaller or comparable number of parameters. 

We observe two interesting patterns for \texttt{Seed-X-PPO-7B}: 1) Its US split BLEU is much higher than the UK and China splits, whereas BLEU differences across locations are smaller for the other models. 2) We find a moderate-to-strong negative correlation between BLEU and total error count for Chinese, with \texttt{Seed-X-PPO-7B} being an outlier---Pearson's $r_\text{incl}\!=\!-0.64$ and $r_\text{excl}=-0.86$ when including and excluding it. Despite having an error count comparable to \texttt{Qwen3.5-9B} and higher than larger models, it obtains a substantially higher BLEU than all other models in \Cref{tab:full-error-table}. This mismatch between error counts and BLEU hints at the possibility of FLORES-like data contamination, whereby overfitting to domain, style, or sentence structure may inflate automatic scores despite the presence of errors.

Broadly, our observed error patterns in model size and locale difficulty, despite a small scale, are consistent with, and provide qualitative support for, our earlier findings based on BLEU, chrF, and $\Delta$ scores.

\subsection{Locale-specific error patterns}

Locale-specific errors substantially outnumber general translation errors. In aggregate, named entities (NE) errors are the most frequent, accounting for $>\!80\%$ of locale-specific errors and $>\!50\%$ of all recorded errors. This pattern is consistent across models and locations, indicating that the accurate translation of localised entities remains a primary source of errors even when the general translation is successful.

The remaining locale-specific error types exhibit more pronounced regional variation. Errors involving idioms and culturally specific expressions (IC) are more frequent in the China split. As expected, Pinyin (PY) transcription errors are exclusive to China, making up 20\% of all errors. These errors arise when Chinese names require romanisation rather than translation. The asymmetry with the UK and US splits is consequential: names in those splits originate from English and are represented in Chinese before being translated back into English, whereas the China split requires the hard-to-generalise Chinese-to-Latin script mapping ability. Thus, the China split presents a unique difficulty of script- and language-specific entity transformation that is absent from other localisations.

Number and unit errors (NU) also show substantial regional variation. The UK and China splits contain 22 and 14 such errors, respectively, compared with only 5 in the US split. These errors involve region-specific measurements and formatting, such as jin (weight), British pounds (weight and currency), and chronological conversions (century and decade to year). The concentration of such errors in the China and UK splits suggests that models are more likely to struggle when the source text contains numbers or units that differ from those in general-purpose translation data.

Overall, the error analysis indicates that localisation affects the difficulty of translation, reflected in the nature of the errors that models make, which vary for different locales. This underscores the value of a locale-oriented test set with a unique source of locale-specific difficulties. 

\section{Conclusion}
This work introduces \ours{}, a source-contrastive, locale-oriented test set that complements the current massive multilingual translation evaluation. \ours{} isolates a model's robustness to cultural content from general translation capability and serves as a diagnostic tool for data contamination. Evaluating 32 models reveals that smaller and MT-specialised systems perform worse on localised content. Furthermore, our contrastive design exposes potential FLORES overfitting and a pervasive US-centric bias, where models translate US-grounded content better than content from native regions. Ultimately, \ours{} demonstrates that true multilingual proficiency requires locale-aware evaluation to accurately reflect real-world use cases.

\section*{Acknowledgements}
AI assistants have been used for coding, visualisations, and paper editing. Conceptualisation, methodology, experiments and analysis, and writing were done by the authors. 

We thank Teresa Sy Ortin and Bouazza Laracha for data contribution. We are grateful for the computing resources from the Northern Ireland High Performance Computing (NI-HPC) service funded by EPSRC (EP/T022175).

\bibliography{custom}

\appendix

\begin{figure*}[th]
\section{LLM Localisation Prompts}
\label{sec:LLM Localisation Prompts}
\vspace{2ex}
\centering\small
\input{figures/llm_paraphrase_prompt}
\caption{System and user prompts used for LLM localisation}
\label{fig:System and user prompts used for LLM localisation}
\end{figure*}

\begin{figure*}[th]
\section{Instructions Presented to Annotators}
\label{sec:Instructions Presented to Annotators}
\vspace{2ex}
\centering\small
\input{figures/annotation_instruction_given_to_humans}
\caption{Annotation instruction given to human annotators}
\label{fig:Project details and annotation instruction given to human annotators}
\end{figure*}

\begin{figure*}[t]
\section{\texorpdfstring{$\Delta$}{} Breakdown by Model and Locale}
\label{sec:delta breakdown}
\vspace{1ex}
\centering\small
\resizebox{0.8\textwidth}{!}{
\begin{tikzpicture}
\begin{axis}[
    scale only axis,
    enlargelimits=false,
    xtick={0,1,2,3,4,5,6,7,8,9,10,11,12,13,14,15,16,17,18,19,20,21,22,23,24,25,26},
    ytick={0,1,2,3,4,5,6,7,8,9,10,11,12,13,14,15,16,17,18,19,20,21,22,23,24,25,26,27,28,29,30,31},
    xticklabels={plt\_Latn\_Madagascar,gom\_Deva\_India,yor\_Latn\_Nigeria,hin\_Deva\_US,tel\_Telu\_India,hin\_Deva\_India,ben\_Beng\_India,cmn\_Hans\_US,pol\_Latn\_Poland,spa\_Latn\_Spain,deu\_Latn\_Germany,bul\_Cyrl\_Bulgaria,hin\_Deva\_Germany,slk\_Latn\_Slovakia,fil\_Latn\_Philippines,tur\_Latn\_Turkey,cmn\_Hans\_UK,cat\_Latn\_Spain,ces\_Latn\_Czechia,cmn\_Hans\_China,rus\_Cyrl\_Russia,deu\_Latn\_Switzerland,acm\_Arab\_Iraq,khk\_Cyrl\_Mongolia,jpn\_Jpan\_Japan,apc\_Arab\_Syria,cmn\_Hans\_Singapore},
    yticklabels={Qwen3.5-35B-A3B,Llama-3.1-8B-instruct,Ministral-3-8B-Instruct-2512,EuroLLM-9B-instruct-2512,Phi-4-mini-instruct,gemma-3-4b-it,Qwen3.5-122B-A10B-FP8,Llama-3.2-3B-instruct,Qwen3.5-27B,Olmo-3-7B-Instruct,Qwen3.5-9B,Qwen3.5-2B,Ministral-3-3B-Instruct-2512,EuroLLM-1.7B-Instruct,Qwen3.5-4B,Llama-3.2-1B-instruct,gemma-3-1b-it,tiny-aya-global,SmolLM-1.7B-Instruct,emma-500-llama3.1-8b-mono,emma-500-llama3.1-8b-bi,madlad400-7b-mt,Llama-3.3-70B-Instruct,Qwen3.5-0.8B,nllb-200-distilled-1.3B,madlad400-3b-mt,nllb-200-distilled-600M,Hy-MT2-1.8B,Hy-MT2-7B,Hy-MT2-30B-A3B,aya-expanse-8b,Seed-X-PPO-7B},
    xmin=-0.5,
    xmax=26.5,
    ymin=-0.5,
    ymax=31.5,
    xticklabel style={rotate=50, anchor=east, font=\tiny,scale=0.85,xshift=1pt,yshift=-2.5pt},
    yticklabel style={rotate=35, anchor=east, font=\tiny,scale=0.85,xshift=2pt,yshift=3pt},
    axis line style={draw=none},
    tick style={draw=none},
    y dir=reverse,
    colorbar horizontal, 
    colorbar style={
        x dir=reverse, 
        height=0.15cm, 
        xticklabel style={font=\tiny}, 
        yshift=9.65cm,
        title={$\Delta_\text{BLEU}$},
        title style={
            at={(0, -1)}, 
            anchor=east, 
            font=\footnotesize,
            xshift=-1pt
        }
    },
    colormap={bright_gold_purple_heatmap}{
        rgb255(0)=(205,50,120);  
        rgb255(1)=(177,61,144);
        rgb255(2)=(149,72,167);
        rgb255(3)=(121,83,191);  
        rgb255(4)=(100,103,196);
        rgb255(5)=(89,138,173);
        rgb255(6)=(78,172,150);
        rgb255(7)=(71,205,125);  
        rgb255(8)=(163,203,56);
        rgb255(9)=(238,198,5);
        rgb255(10)=(238,180,34); 
    },
    point meta min=-18.5969,
    point meta max=8.1575,
]
\addplot[
    matrix plot*,
    point meta=explicit,
    mesh/cols=27
] table[x=language_idx, y=model_idx, meta=delta_bleu] {data/heatmap_delta_bleu_sorted.tsv};
\end{axis}
\end{tikzpicture}
}
\resizebox{0.8\textwidth}{!}{
\begin{tikzpicture}
\begin{axis}[
    scale only axis,
    enlargelimits=false,
    xtick={0,1,2,3,4,5,6,7,8,9,10,11,12,13,14,15,16,17,18,19,20,21,22,23,24,25,26},
    ytick={0,1,2,3,4,5,6,7,8,9,10,11,12,13,14,15,16,17,18,19,20,21,22,23,24,25,26,27,28,29,30,31},
    xticklabels={hin\_Deva\_US,cmn\_Hans\_US,plt\_Latn\_Madagascar,yor\_Latn\_Nigeria,gom\_Deva\_India,tel\_Telu\_India,ben\_Beng\_India,cmn\_Hans\_UK,fil\_Latn\_Philippines,bul\_Cyrl\_Bulgaria,hin\_Deva\_India,spa\_Latn\_Spain,cmn\_Hans\_China,deu\_Latn\_Germany,cat\_Latn\_Spain,hin\_Deva\_Germany,slk\_Latn\_Slovakia,rus\_Cyrl\_Russia,pol\_Latn\_Poland,tur\_Latn\_Turkey,ces\_Latn\_Czechia,cmn\_Hans\_Singapore,deu\_Latn\_Switzerland,acm\_Arab\_Iraq,apc\_Arab\_Syria,jpn\_Jpan\_Japan,khk\_Cyrl\_Mongolia},
    yticklabels={Llama-3.1-8B-instruct,Qwen3.5-35B-A3B,SmolLM-1.7B-Instruct,EuroLLM-9B-instruct-2512,Ministral-3-8B-Instruct-2512,Qwen3.5-122B-A10B-FP8,gemma-3-4b-it,Qwen3.5-27B,Llama-3.2-3B-instruct,Qwen3.5-9B,Ministral-3-3B-Instruct-2512,Qwen3.5-2B,EuroLLM-1.7B-Instruct,Olmo-3-7B-Instruct,Phi-4-mini-instruct,Qwen3.5-4B,emma-500-llama3.1-8b-bi,gemma-3-1b-it,tiny-aya-global,Llama-3.2-1B-instruct,emma-500-llama3.1-8b-mono,Llama-3.3-70B-Instruct,madlad400-7b-mt,Qwen3.5-0.8B,nllb-200-distilled-1.3B,madlad400-3b-mt,nllb-200-distilled-600M,Hy-MT2-7B,Hy-MT2-1.8B,Hy-MT2-30B-A3B,aya-expanse-8b,Seed-X-PPO-7B},
    xmin=-0.5,
    xmax=26.5,
    ymin=-0.5,
    ymax=31.5,
    xticklabel style={rotate=50, anchor=east, font=\tiny,scale=0.85,xshift=1pt,yshift=-2.5pt},
    yticklabel style={rotate=35, anchor=east, font=\tiny,scale=0.85,xshift=2pt,yshift=3pt},
    axis line style={draw=none},
    tick style={draw=none},
    y dir=reverse,
    colorbar horizontal,
    colorbar style={
        x dir=reverse,
        height=0.15cm,
        xticklabel style={font=\tiny},
        yshift=0.3cm,
        title={$\Delta_\text{chrF}$},
        title style={
            at={(0, -1)},
            anchor=east,
            font=\footnotesize,
            xshift=-1pt
        }
    },
    colormap={bright_gold_purple_heatmap}{
        rgb255(0)=(205,50,120);
        rgb255(1)=(177,61,144);
        rgb255(2)=(149,72,167);
        rgb255(3)=(121,83,191);
        rgb255(4)=(100,103,196);
        rgb255(5)=(89,138,173);
        rgb255(6)=(78,172,150);
        rgb255(7)=(71,205,125);
        rgb255(8)=(163,203,56);
        rgb255(9)=(238,198,5);
        rgb255(10)=(238,180,34);
    },
    point meta min=-11.4534,
    point meta max=4.4672,
]
\addplot[
    matrix plot*,
    point meta=explicit,
    mesh/cols=27
] table[x=language_idx, y=model_idx, meta=delta_chrf] {data/heatmap_delta_chrf_sorted.tsv};
\end{axis}
\end{tikzpicture}
}
\caption{Heatmaps of $\Delta_\text{BLEU}$ (top) and $\Delta_\text{chrF}$ (bottom). Models (y-axis) are sorted by mean $\Delta$ across locales, and locales (x-axis) are sorted by mean $\Delta$ across models, with more negative $\Delta$ towards the bottom and right.}
\label{fig:breakdown of deltas}
\end{figure*}
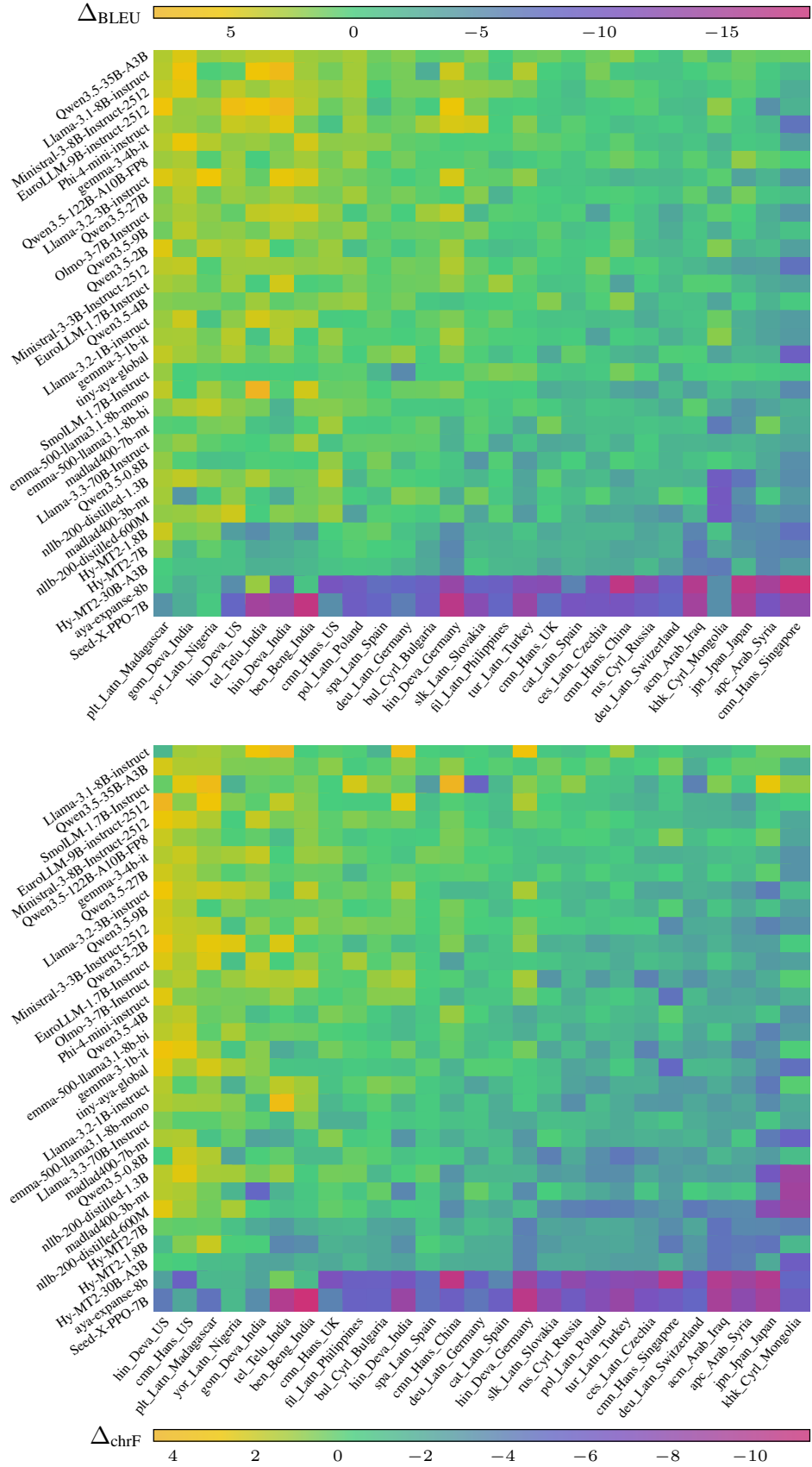

\end{document}